\documentclass[11pt,a4paper]{article}
\usepackage[hyperref]{acl2021}
\usepackage{times}
\usepackage{latexsym}
\usepackage{soul}

\usepackage{dirtytalk}
\usepackage{multicol}
\usepackage{threeparttable}  
\usepackage{multirow}
\usepackage{makecell}
\usepackage{array}
\usepackage{amssymb}
\usepackage{cuted} 
\usepackage{amsmath}
\usepackage{afterpage}  
\usepackage{subcaption}
\usepackage{float}
\usepackage{microtype}
\newcommand{\cmmnt}[1]{}

\aclfinalcopy    

\usepackage{colortbl}
\usepackage{graphicx}
\usepackage{diagbox}
\usepackage{booktabs} 
\usepackage{tablefootnote}

\title{From Image Captioning to Visual Storytelling}
\author{Admitos Passadakis \\
  \texttt{apassadakis@tudelft.nl} \\\And
  Yingjin Song \\
  \texttt{y.song5@uu.nl} \\\And
  Albert Gatt \\ 
  \texttt{a.gatt@uu.nl}
  }
\begin{document}
\maketitle

\begin{abstract}
Visual Storytelling is a challenging multimodal task between Vision \& Language, where the purpose is to generate a story for a stream of  images. 
Its difficulty lies on the fact that the story should be both grounded to the image sequence but also narrative and coherent. The aim of this work is to balance between these aspects, by treating Visual Storytelling as a superset of Image Captioning, an approach quite different compared to most of prior relevant studies.   
This means that we firstly employ a vision-to-language model for obtaining captions of the input images, and then, these captions are transformed into coherent narratives using language-to-language methods.
Our multifarious evaluation shows that integrating  captioning and storytelling under a unified framework, has a positive impact on the quality of the produced stories.
In addition, compared to numerous previous studies, this approach accelerates training time and makes our framework readily reusable and reproducible by anyone interested.
Lastly, we propose a new metric/tool, named \textit{ideality}, that can be used to simulate how far some results are from an oracle model, and we apply it to emulate human-likeness in visual storytelling. 
\end{abstract}


\section{Introduction}
The task of generating a textual story given a series of input images that are potentially correlated, is called \textit{Visual Storytelling} \cite{huang-etal-2016-visual}. This is a challenging multimodal task that lies in the intersection of Computer Vision (CV) and Natural Language Processing (NLP). 
In most cases, Visual Storytelling is regarded to be a distinct task from \textit{Image Captioning} \cite{DBLP:journals/corr/LinMBHPRDZ14}, which focuses on describing objects and their attributes. 
Indeed, Visual Storytelling is not only about mere image descriptions, as it aims to bridge the semantic gap between visual data and textual representations by introducing a narrative form, coherence and a sense of chronological progress in the output text.

While Image Captioning models have performed well in generating detailed descriptions of individual images \cite{anderson2018bottomuptopdownattentionimage,lu2019vilbertpretrainingtaskagnosticvisiolinguistic}, storytelling necessitates more sensitivity to temporal relationships and more narrative structure. 
Existing visual storytelling methods have mainly focused on end-to-end systems, which map images sequences directly into stories \cite{10.5555/3504035.3504941, Smilevski_2018, kim2019glacnetglocalattention, su2020berthlstmsberthierarchicallstms, jung2020hide, Yu_2021_CVPR}, by utilizing typical visual-encoder \& text-decoder architectures.
While these approaches have been promising regarding \textit{visual grounding}, they often fall short in resolving two other crucial components of storytelling: \textit{narrative coherence} and \textit{semantic richness} \cite{10.55556/3298239.3298450,NIPS2015_17e62166}. 
The issue of joint optimization of these dimensions has given rise to various solutions, from knowledge-enriched techniques \cite{hsu2019knowledgeenrichedvisualstorytelling, hsu2021plotreworkmodelingstorylines, chen2021commonsenseknowledgeawareconcept} to more recent applications of multi-modal \textit{Large Language Models (LLMs)} \cite{liu2023visualinstructiontuning, zhang2024visionlanguagemodelsvisiontasks}.

Here, we propose an alternative to this challenge by addressing Visual Storytelling formally as a superset of Image Captioning. 
Our solution splits storytelling into two separate stages: first, we create precise and expressive descriptions of single images with an advanced captioning model, and then, we convert these descriptions into coherent stories using sophisticated language-to-language methods. 
This decomposition has several advantages: it leverages the maturity of image captioning models, simplifies storytelling by separating visual understanding from narrative generation, and makes the overall system more reproducible\footnote{In fact, our system can be tested online by clicking \href{https://github.com/arpajj/Visual-Storyteller}{HERE}. Please, notify the first author before proceeding to usage.}.

Another challenge in visual storytelling is that there is no objectively correct and acceptable way of measuring which story can be considered as complete. Despite that \textit{visual grounding} is a commonly admissible dimension, the same cannot be said for \textit{narrative coherence} and \textit{semantic richness}. 
Different studies have attempted to capture all these metrics, by recruiting different methods, algorithms or language models \cite{wang2022rovistlearningrobustmetricsvisual, surikuchi2023groovistmetricgroundingobjects}.
To overcome this hurdle, we additionally propose a novel metric that can be utilized in conjunction with other metrics, that measure similar characteristics, and with one oracle method, such as human writing, that is objectively accepted as a comprehensive way of written text. 
As a result, we believe that this new metric could simulate some aspects of human-likeness in the generated stories.  
Having said these, we can summarize the main contributions of this research as follows: 
\vspace{-0.2cm}
\begin{itemize}
    \item We propose a simple, yet very effective visual storytelling method which is lightweight, easily reusable and reproducible. Its efficiency is shown through multiple evaluation methods. \vspace{-0.15cm}
    \item We successfully unify two different Vision \& Language tasks, Image Captioning and Visual Storytelling, under a single-stream pipeline. 
    \vspace{-0.15cm}
    \item We also propose a new metric/tool that can be universally used in order to measure how close we are to an ideal result. In our case, we utilize it to simulate how human-like are our generated stories in linguistic terms.  
\end{itemize}


\section{Related Work}
\subsection{Image Captioning}
Image captioning has evolved from early hand-tuned approaches based on manually crafted features and linguistic rules \cite{farhadi2010every,devlin2015exploringnearestneighborapproaches} to modern deep learning methods like CNNs for feature extraction and RNNs for sequential text generation. 
The introduction of Microsoft COCO dataset \cite{DBLP:journals/corr/LinMBHPRDZ14}, provided large-scale, diverse image-caption pairs, allowing standard evaluation metrics to be set and models to be benchmarked \cite{vinyals2015tellneuralimagecaption}.  
Subsequent advancements introduced attention mechanisms which enabled models to focus on the appropriate regions of an image \cite{xu2016showattendtellneural,lu2017knowinglookadaptiveattention}. Reinforcement Learning (RL) for optimizing captioning models was added in subsequent enhancements \cite{rennie2017selfcriticalsequencetrainingimage}, whilst graph-based approaches utilized scene graphs to facilitate contextual understanding \cite{yang2018autoencodingscenegraphsimage,yao2018exploringvisualrelationshipimage}. 
The transition to transformers-based methods also introduced new improvements, such as the attention-on-attention networks \cite{huang2019attentionattentionimagecaptioning} and entangled spatial-aware Transformers \cite{herdade2020imagecaptioningtransformingobjects, he2020imagecaptioningimagetransformer}. 

\subsection{Visual Storytelling}
Visual Storytelling (VIST) \cite{huang-etal-2016-visual} is a challenging multimodal task that aims to generate coherent and human-like narratives from sequences of ordered images and it is often  differentiated from Image Captioning.
Initial attempts built upon image captioning models, incorporating global-local visual attention \cite{kim2019glacnetglocalattention} and contextualized image representations \cite{gonzalezrico2018contextualizetellneuralvisual} to improve narrative fluency. 
Traditional methods relied on CNNs and LSTMs \cite{kim2015understanding, yu2017hierarchicallyattentivernnalbumsummarization, Yu_2021_CVPR}, while more recent approaches leverage Transformer architectures to integrate image and text features \cite{10.1145/3474085.3475236,10.1145/3469877.3490604}. 
RL was also used for VIST, with custom reward functions guiding story generation \cite{10.5555/3504035.3504941, wang2018metricsperfectadversarialreward, huang2019attentionattentionimagecaptioning, hu2020makesgoodstorydesigning, chen2024tarnvisttopicawarereinforcement}, though its training instability remains a challenge. 
Beyond data-driven models, external knowledge integration played a crucial role in enhancing story richness and logical coherence, enabling models to reason about commonsense concepts that are not explicitly present in images \cite{yang2019knowledgeable, hsu2019knowledgeenrichedvisualstorytelling,xu2021imagine,chen2021commonsenseknowledgeawareconcept, wang2024scovistsocialinteractioncommonsense}.

A key limitation in VIST is the scarcity of high-quality features, which led to the introduction of other similar datasets, such as AESOP \cite{Ravi_2021_ICCV}, which features synthetic image sequences with enriched textual descriptions, and VWP \cite{hong2023visualwritingpromptscharactergrounded} that curates movie frame sequences to improve storytelling quality. 
To further enhance narrative generation, some models employ semantic scene graphs and Graph Convolutional Networks to capture fine-grained object relations \cite{han2020victrvisualinformationcaptured, hong-etal-2020-diverse,Wang_Wei_Li_Zhang_Huang_2020}. 
Various approaches also consider other storytelling-specific factors like sentiments \cite{9797749} or major characters \cite{prabhumoye2019mywaytellingstory,liu2023detectinggroundingimportantcharacters,liu2024generatingvisualstoriesgrounded}.

\subsection{Text-to-Text reformulation}
Recent advancements in text-to-text story generation focus on coherence, controllability and knowledge integration. Early attempts used LSTM-based sequence-to-sequence models, along with Statistical Machine Translation approaches or plan-and-write techniques \cite{jain2017storygenerationsequenceindependent, yao2019planandwritebetterautomaticstorytelling}.  
Later studies, leveraged transformers models enhanced by commonsense knowledge, multi-task learning and external knowledge for controlled generation \cite{guan2020knowledgeenhancedpretrainingmodelcommonsense,xu-etal-2020-megatron}. 
Other works, in order to improve fluency, consistency and coherence in storytelling, attempted to mix methods like \textit{Variational Autoencoders} and transformers with latent representations \cite{fang2021transformerbasedconditionalvariationalautoencoder}. 
Lastly, very similar to our methodology, some approaches have incorporated such text-to-text manipulations within the visual storytelling task itself \cite{yang-jin-2023-attractive,10.1145/3581783.3612179}. 


\section{Methods}
\label{sec:methods}

\subsection{Dataset}
\label{subsec:dataset}
We utilize the Visual Storytelling dataset - VIST \cite{huang-etal-2016-visual}, a well-known dataset for sequential vision-to-language tasks. 
It consists of 10,049 Flickr albums with 209,652 unique photos and its textual information is divided in three tiers: 1) \textit{Descriptions of images-in-isolation (DII)}; 2) \textit{Descriptions of images-in-sequence (DIS)}; and 3) \textit{Stories for images-in-sequence (SIS)}.
For our purpose, we make use of the annotations in the DII and SIS tiers\footnote{These two tiers don not have they same size, with SIS being much larger than DII.}.
The analytical statistics of VIST dataset regarding the number of albums and the story-samples are presented on Table \ref{tab:vist_stats}. 
From there, each story-sample contains an image-sequence of 5 images from a photo album that is described by 1 story (usually one sentence per image).
Given the focus of this work, we utilize the portion of SIS for which the corresponding DIS captions are available. 
Moreover, some internal miss-structures of the dataset do exist, such as broken images. Even if one image is broken, then the whole story-sequence is discarded.  
Due to these factors, the total number of story-samples in training, validation and test sets is diminished even further (see Table \ref{tab:vist_stats}).

After applying these filtering steps and assuming that our dataset has length of $N$, the data is organized on aligned triplets: $T_i = \{X_i, C_i, S_i\}_{i=1}^N$. 
Here, $X_i$ is the set of images, $C_i$ is the set of captions corresponding to the images and $S_i$ is the set of story-sentences also corresponding to the captions and images. 
In most cases, the common number of elements in all sets $X_i$, $C_i$, and $S_i$ is $M=5$\footnote{However, for the storytelling phase, since BART accepts inputs as concatenated text sequences fewer/more captions/story-sentences can be handled. In fact, this  makes our framework versatile in producing stories with more/less than $5$ sentences in many occasions.}.
Additionally, we represent with $x_i^m \in X_i$ the $m^{th}$ image of the set $X_i$, and with $c_i^m \in C_i$ we represent the $m^{th}$ caption of the set $C_i$. 
Also, each caption can be seen as a sequence of tokens $c_i^m = c_{i,1}^m, \ldots, c_{i,\ell}^m$, where tokens have been padded to a maximal length $\ell$.
Respectively, for the storytelling phase, we assume that $s_i^m \in S_i$ represents the $m^{th}$ sentence of the story-set $S_i$. 

\begin{table}[ht]
\centering
\small
\begin{tabular}{lccc}
\toprule
\textbf{Split} & \makecell{\textbf{Albums}} & \makecell{\textbf{Story-}\\\textbf{samples}} & \makecell{\textbf{Filtered}\\\textbf{Story-samples}} \\
\midrule
Train      & 8,037 & 40,181 & 26,570 \\
Validation &   999 &  4,998 &  3,319 \\
Test       & 1,013 &  5,066 &  3,338 \\
\midrule
\textbf{Total} & \textbf{10,049} & \textbf{50,245} & \textbf{33,227} \\
\bottomrule
\end{tabular}
\caption{ \centering Statistics of VIST dataset across splits.}
\label{tab:vist_stats}
\end{table}

\vspace{-0.1cm}
\subsection{The Framework}
\label{subsec:framework}
\begin{figure*}[ht]
    \centering
    \includegraphics[height=3cm, width=0.9\linewidth]{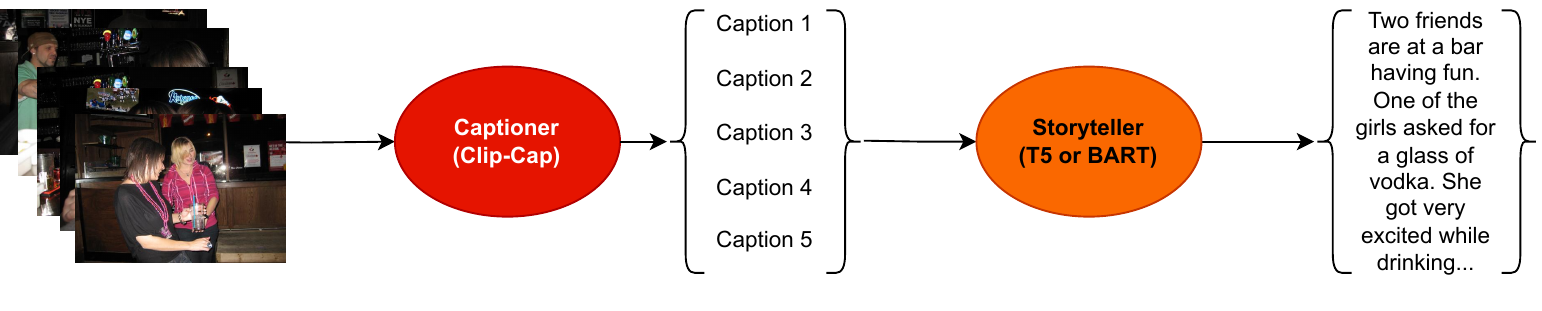}
    \vspace{-0.5cm}
    \caption{ \centering \small {Our proposed framework for Visual Storytelling during testing time, comprises of a captioning and storytelling model.
    Note that the two components are totally independent during training and are only chained on inference time.
    The captioner extracts the central content and information on the input image-sequence and provides descriptions, whilst the storyteller reformulates those captions and adds a taste of narration, coherence and chronicity, traits that eventually make a story.}}
    \label{fig:model_diagram}
\end{figure*}

Our framework essentially comprises of two sub-parts: The \say{Captioner}, a Vision-to-Language model named \textit{ClipCap} \cite{DBLP:journals/corr/abs-2111-09734} and the \say{Storyteller}, which is a Language-to-Language model. 
The former is used for creating captions for the input image sequence, whilst the latter accepts these captions and reformulates them to a more narrative and coherent text (stories). 
An overview of the proposed framework, is given on Fig. \ref{fig:model_diagram}, where firstly, the input image-sequence is given to the captioning model, ClipCap, and then the generated captions are fed to the storyteller model, for which we try two alternative architectures: 1) BART \cite{lewis-etal-2020-bart} 
and 2) T5 \cite{JMLR:v21:20-074}. 
It's worth to note that the two modules are trained separately.

\subsubsection{The Vision-to-Language Model}
\label{sub-subsec:vis-to-lang-model}
ClipCap is a transformer-based network that consists of three parts: a) CLIP \cite{radford2021learning}, b) a Mapping Network and c) a Language Generator. 
CLIP is a multimodal vision-language model that is trained using contrastive learning, associating similar image-text pairs and differentiating the dissimilar ones.
The key component of ClipCap is the mapping network, which
translates the CLIP embeddings to the generator's input space. Depending on whether this language generator is fine-tuned, two different mapping networks are used. In the first case, a simple MLP is deployed, whilst in the second, a more expressive transformer is utilized.
Conversely to Mokady et al., in our experiment we use both types of mapping networks, while always fine-tuning the generator. This generator model is GPT2 \cite{radford2019language}.

The first operation of ClipCap is to encode each image $x_i^m$ by using the visual encoder of the pre-trained CLIP model. Next, a light mapping network, denoted as $F$, is employed to map the CLIP visual embedding to $k$ embedding vectors:
\vspace{-0.3em}
\begin{equation}
p_{i,1}^m, \ldots, p_{i,k}^m = F(\text{CLIP}(x_i^m)).
\end{equation}
Each vector $p_{i,j}^m$ is essentially a visual element embedding and has the same dimension as a word embedding. Then, the projected visual embedding is concatenated with the caption embeddings, $c_i^m$:
\begin{equation}
Z^i = [p_i^m ; c_i^m] = [p_{i,1}^m, \ldots, p_{i,k}^m ; c_{i,1}^m, \ldots, c_{i,\ell}^m].
\end{equation}
Ultimately, the language model takes the concatenated prefix-caption  pair $\{ Z^i \}_{i=1}^N$, and learns to predict the caption tokens autoregressively, conditioned on the image-prefix. As a result, both itself and the  mapping network $F$ are trained with cross-entropy loss as:
{\small
\begin{equation}
\mathcal{L}_1 = - \sum_{i=1}^N \sum_{m=1}^M \sum_{j=1}^\ell 
    \log p_\theta \big(c_{i,j}^m \,|\, p_i^m ; c_{i,1}^m, \ldots, c_{i,j-1}^m \big).
\end{equation}
}


\subsubsection{The Language-to-Language Model}
\label{sub-subsec:lang-to-lang-model}
\vspace{-0.1cm}
Our two options for language-to-language generation were BART and T5. BART \cite{lewis-etal-2020-bart} is a denoising autoencoder transformer trained to reconstruct corrupted input using techniques like \textit{token masking} and \textit{deletion}. It combines characteristics of both encoder-based \cite{devlin-etal-2019-bert} and decoder-based \cite{radford2018improving} transformers within a standard encoder-decoder architecture \cite{DBLP:journals/corr/VaswaniSPUJGKP17}. Similarly, T5  \cite{JMLR:v21:20-074} is also an encoder-decoder model, pre-trained using a generalized \textit{masked language modeling} approach known as \textit{span corruption}.

\begin{figure}[ht]
    \hspace{-0.66cm}\includegraphics[height=9cm, width=1.15\linewidth]{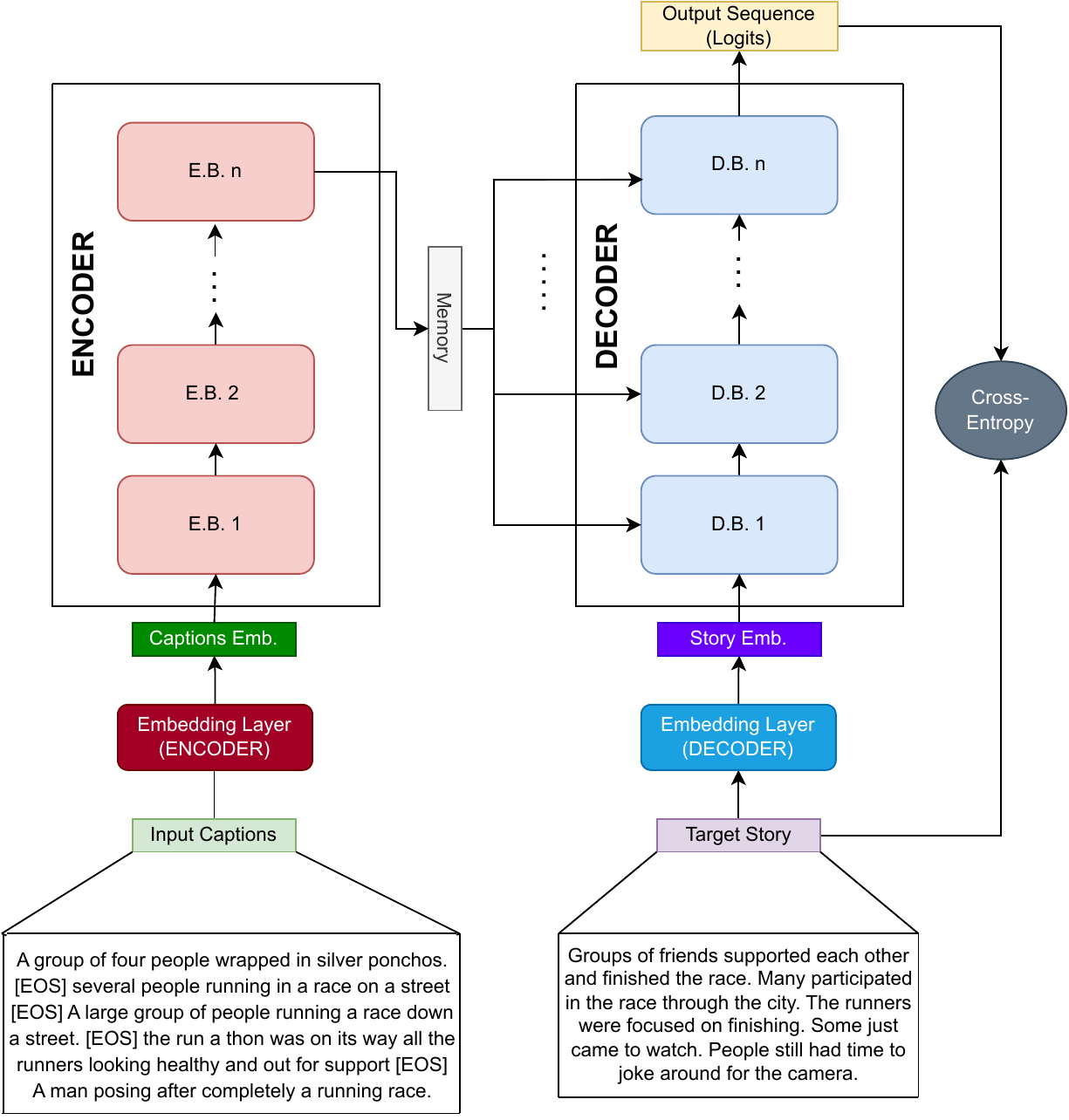}
    \vspace{-0.5cm}
    \caption{ \centering \small {Storyteller's architecture (T5 or BART), which is utilized for reformulating the captions of VIST to coherent narratives. During training, the captions embeddings are given to the stack of encoders, while the story embeddings are fed to the stack of decoders. In addition, the top head's output of the encoder (memory) is passed to the decoder for keeping the reconstructed story grounded to the input. The decoder learns to generate storylines by using Cross-Entropy loss.}}
    \label{fig:storyteller}
    \vspace{-0.2cm}
\end{figure}

In order to adapt these two models in to our storytelling setting, we had to separately use their encoder and decoder components of the architecture for the captions and the story reconstruction correspondingly. 
For this reason, we denote their encoder as $\mathbf{E}$, its decoder as $\mathbf{D}$.
Firstly, all the captions $c_i^m$ and story-sentences $s_i^m$ are concatenated with each other as following: 
\vspace{-0.1cm}
\begin{subequations}
\begin{align}
\Tilde{C_i} &= [c_{i}^1 ; [eos] ; c_{i}^2 [eos] ; \ldots ; [eos] ; c_{i}^M], \\ 
\Tilde{S_i} &= [s_{i}^1 ; s_{i}^2 ; \ldots ; s_{i}^M].
\vspace{-0.3cm}
\end{align}
\label{eq:concat}
\end{subequations}
In equations (\ref{eq:concat}), $\Tilde{C_i}$ and $\Tilde{S_i}$ are the concatenated caption-sets and story-sets, which are inputted to $\mathbf{E}$ and $\mathbf{D}$ respectively. 
Nevertheless, in the case of the captions, in order for making the story generator able to capture the contextual difference between each caption the concatenation is made using the $[eos]$ token. 
In addition, the concatenated story-set $\Tilde{S_i}$, essentially consists a story-sequence of tokens $\Tilde{S_i} = s_{i,1}, s_{i,2}, \ldots, s_{i,n}$, where each story has been padded to a maximal length $n$.
Therefore, the reconstructed story, $\hat{S_i}$, by the text-to-text model during training is given by the equation:
\vspace{-0.1cm}
\begin{equation}
\hat{S_i} =  \mathbf{D}(\Tilde{S_i},\mathbf{E}(\Tilde{C_i})),
\end{equation}
where $\mathbf{E}(\Tilde{C_i})$ is the injected memory from the encoder to all decoder's blocks. 
Similarly to ClipCap, where the language model learns to predict captions autoregressively, BART's or T5's decoder learns to generate the next word in the story, conditioned on the previously generated content and the captions. 
This entails that our storyteller is also trained with cross-entropy loss function:
\vspace{-0.1cm}
\begin{equation}
\mathcal{L}_2 = - \sum_{i=1}^N \sum_{j=1}^n 
    \log p_\theta \big(s_{i,j} | \Tilde{C}_i , s_{i,1}, \ldots, s_{i,j-1} \big).
\end{equation}
A schematic representation of how our storyteller is trained is depicted on Fig. \ref{fig:storyteller}.

\subsection{Data Augmentation - Revolving Training}
\label{subsec:data_aug}
The most challenging part from the two phases is the storytelling stage, since there are no exact guidelines of what is consider to be a good story \cite{hu2020makesgoodstorydesigning}. 
Plenty of prior works attempts to focus on cohesion, visual relevance, general concept knowledge and other traits for creating complete stories. In order to do so, they exploit several powerful but not easily reproducible techniques such as RL or knowledge graphs \cite{wang2018metricsperfectadversarialreward, hsu2019knowledgeenrichedvisualstorytelling, xu2021imagine}. 

In our work, for dealing with this challenge and making our framework more robust, we simply exploit the different type of descriptions that VIST dataset offers. 
More specifically, the caption-set $C_i$, is not comprised by $M$ captions, but  essentially from $M$ lists, each of which contains $3$ possible captions for the corresponding image.
Hence, a more accurate representation of a data triplet of VIST, would be: $T_i = \{X_i, \{C_{i,j}\}_{j=1}^3, S_i\}_{i=1}^N$.
To that end, we leverage these three captions to train our captioner and storyteller on three slightly different sub-datasets consisting of triplets of: $T_{1_i} = \{X_i, C_{i,1}, S_i \}_{i=1}^N$, $T_{2_i} = \{X_i, C_{i,2}, S_i \}_{i=1}^N$ and $T_{3_i} = \{X_i, C_{i,3}, S_i \}_{i=1}^N$.  
Therefore, we significantly augment the data size and we are capable of training our models in a revolving manner between $T_{1}$, $T_{2}$ and $T_{3}$.  
During this revolving training, both the captioner and the storyteller are trained on each sub-dataset for one third of the total epochs.  

\section{Experimental Set-up}
\label{sec:set_up}
\subsection{Choice of components}
\label{subsec:choice}
We use two pre-trained versions of ClipCap captioner from Mokady et al., and we further fine-tune them on VIST dataset. 
The first option is ClipCap with an MLP mapping network and the second exploits a Transformer type mapper. 
For the former, the image encoder of CLIP is a $ViT_{Huge}$ \cite{dosovitskiy2021imageworth16x16words}, which transforms input images to an embedding with size $512$, whereas for the latter 
the CLIP image encoder is a ResNet-$50$ \cite{he2015deepresiduallearningimage}, with an image-embedding of $640$. 

As a result, the condensed visual embedding (prefix) for concatenation with the captions has a size of $10$ and $40$ for the MLP and the Transformer case correspondingly. 
As we already underscored, for the captioning phase, we always fine-tune GPT2 (small) and we use its tokenizer to project the captions to a sequence of embeddings regarding data subset $T_{A_i}$. 
For the story phase, BART (large) and T5 (base) with their own tokenizer, project both the captions and the story-sentences to embedding sequences regarding the data subset $T_{B_i}$.

\subsection{Training procedure \& Model size}
As already underlined, each part of our framework is fine-tuned separately on the corresponding train part of VIST dataset (and the validation part respectively). 
This means that for our captioner we use the \say{image-to-caption} subset of initial triplet, that is: $T_{A_i} = \{X_i, \{C_{i,j}\}_{j=1}^3\}_{i=1}^N$, while for the storyteller we use \say{caption-to-story} part from the initial format: $T_{B_i} = \{\{C_{i,j}\}_{j=1}^3, S_i\}_{i=1}^N$. 
This totally independent fine-tuning of the 2 sub-parts, minimizes training time (in practice, it can be carried out in parallel) and makes our framework easily reusable and reproducible. 

We train both versions of the captioning (using an MLP or Transformer as mapping network) and the storytelling (placing T5 or BART as generators) models for $15$ epochs, which means that for each of the three versions of their respective sub-datasets\footnote{$T_{A1}$, $T_{A2}$, $T_{A3}$ and $T_{B1}$, $T_{B2}$, $T_{B3}$}, they get $5$ epochs. 
Moreover, for both networks we opt batch size of $16$ and  
AdamW \cite{kingma2017adammethodstochasticoptimization} with weight decay fix \cite{loshchilov2019decoupledweightdecayregularization} as an optimization algorithm with a learning rate of $4e^{-5}$ and $1000$ warm-up steps.
It should be noted that all our experiments are conducted on an RTX-2080 GPU.
Lastly, we created four distinct frameworks for the test phase, after combining the two types of mapping networks in the captioner and the two storytellers that we fine-tuned during training. 
These possible components and the resulted frameworks are given on Table \ref{tab:model_sizes}. Along with each component, we also provide its total number of trainable parameters. 

\begin{table}[t!]
    \centering
    \small
    \begin{tabular}{|>{\raggedright\arraybackslash}m{0.85cm}|m{1.85cm}|m{2.5cm}|}
        \hline
         & \multicolumn{1}{|c|}{\textbf{Model Name}} & \textbf{Trainable Params.} \\ \hline \midrule 
         
         \multirow{2}{*}{\vspace{-0.8cm} \hspace{1.8pt} 
         \rotatebox{90}{\centering \shortstack{\textbf{Captioner}}}} 
        
        & \centering ClipCap - MLP & \hspace{0.55cm} $\sim$$156m$  \\ [1.75ex]
        
        & \centering ClipCap - Transformer & \hspace{0.55cm} $\sim$$182m$ \\
        \midrule
        
        \multirow{2}{*}{\vspace{-0.4cm} \hspace{-1.5pt} 
        \rotatebox{90}{\centering \shortstack{\textbf{Story} \\ \textbf{teller}}}} 
        
        & \centering T5 & \hspace{0.55cm} $\sim$$223m$ \\ [1.4ex]
        & \centering BART & \hspace{0.55cm} $\sim$$406m$ \\  [0.5ex]
        \midrule
        
        \multirow{4}{*}{\vspace{-1.2cm} \hspace{-1.5pt} 
        \rotatebox{90}{\centering \shortstack{\textbf{Assembled} \\ \textbf{Framework}}}} 
        
        & \centering MLP+T5 & \hspace{0.55cm} $\sim$$380m$ \\ [1.2ex]
        & \centering Transformer + T5 & \hspace{0.55cm} $\sim$$405m$ \\  [1.7ex]
        & \centering  MLP+BART & \hspace{0.55cm} $\sim$$560m$ \\ [1.2ex]
        & \centering Transformer + BART & \hspace{0.55cm} $\sim$$600m$ \\  
        \hline
    \end{tabular}
    \caption{\small{Framework's components and trainable parameters}}
    \label{tab:model_sizes}
    \vspace{-0.2cm}
\end{table}

\vspace{-0.2cm}
\subsection{Automatic Evaluation}
\label{subsec:evaluation}
As most of the aforementioned related studies, we evaluate our framework variants on Visual Storytelling task by using automatic metrics. In particular, the most popular Natural Language Generation (NLG) metrics are being utilized. These include BLEU \cite{papineni2002bleu}, METEOR \cite{banerjee-lavie-2005-meteor}, ROUGE\_L \cite{lin2004rouge} and CIDEr \cite{vedantam2015ciderconsensusbasedimagedescription}. 
These metrics are trying to capture n-gram overlap between a candidate and a reference sentence. 

However, Visual Storytelling has been characterized as a complex task, for which standard NLG metrics are not enough for evaluation \cite{wang2018metricsperfectadversarialreward, hsu2019visualstorypostediting}. 
To that end, several alternatives for measuring different aspects of storytelling, such as semantics, relevance with image or story cohesion have been proposed. 
Regarding semantic similarity, here we use SPICE \cite{anderson2016spicesemanticpropositionalimage}, which captures semantic relations between sentences through scene graphs and BLEURT \cite{sellam-etal-2020-bleurt}, which captures semantic similarity by utilizing deep contextual embeddings taken from language models.
Last but not least, for quantifying coherence, visual grounding and language fluency we focalize on two reference-free metrics. These are RoViST \cite{wang2022rovistlearningrobustmetricsvisual} which measures visual grounding, coherence and non-redundancy on the generated stories and Perplexity which measures linguistic fluency.

\subsection{Baseline models}
Since its introduction, Visual Storytelling \cite{huang-etal-2016-visual}, has seen several approaches that consistently improve the state-of-the-art performance. 
In our experimentation we encompass studies ranging back from the early days of VIST till the some of the latest which exploit LLMs.
Namely, we include the following state-of-art baselines: \\
1. \textbf{AREL} \cite{wang2018metricsperfectadversarialreward}: Is an adversarial framework that leverages reinforcement learning to infer an implicit reward function, that guides policy optimization, enabling more effective learning in structured prediction tasks. \\
2. \textbf{GLACNet} \cite{kim2019glacnetglocalattention}: Is a seq2seq model that integrates global and local attention mechanisms with context cascading to enhance visual feature representation. \\
3. \textbf{KG-Story} \cite{hsu2019knowledgeenrichedvisualstorytelling}: Enhances story generation by leveraging external knowledge bases like Visual Genome \cite{krishna2016visualgenomeconnectinglanguage}. Then, constructs conceptual representations for each image, identifies connections between terms, and utilizes a transformer to generate coherent narratives. \\
4. \textbf{ReCo-RL} \cite{hu2020makesgoodstorydesigning}: Is another reinforcement learning model that employs composite rewards to optimize relevance, coherence, and expressiveness in visual storytelling. \\
5. \textbf{PR-VIST} \cite{hsu2021plotreworkmodelingstorylines}: Applies a \say{plot and rework} techique that constructs a story graph by linking nouns with verbs from Visual Genome and VIST datasets. \\
6. \textbf{TAMP} \cite{Yu_2021_CVPR}: Introduces an auxiliary training task to align vision encoder with a pretrained language model using adaptation loss. Also, it enforces temporal cohesion through coherence loss, by aligning text and visual representations. \\
7. \textbf{MCSM+BART} \cite{chen2021commonsenseknowledgeawareconcept}: Enhances story generation by leveraging image concepts and related knowledge from ConceptNet \cite{liu2004conceptnet}. Its Maximal Clique Selection Module learns a correlation map to identify the most relevant concepts, which are then used as input for BART to generate richer narratives. \\
8. \textbf{SCO-VIST} \cite{wang2024scovistsocialinteractioncommonsense}: It models image sequences as a graph, by incorporating objects, relations and knowledge about human actions and social interactions. It constructs a weighted story graph linking plot points through semantic and occurrence-based edges, using the Floyd-Warshall algorithm to generate a coherent storyline. \\
9. \textbf{Ca-VIST} \cite{song2024contextawarevisualstorytellingvisual}: Utilizes visual prefix tuning and contrastive learning for context-aware visual storytelling. It connects vision and language through a lightweight mapping network and leverages pretrained models for better generalization, coherence and visual relevance. \\
10. \textbf{StoryLLaVA} \cite{yang-etal-2025-storyllava}: Is a multimodal LLM that enhances visual storytelling by improving temporal and spatial coherence. It introduces a Topic-Driven Narrative Optimizer (TDNO) to refine training data and the outputs, alongside a preference-based ranking method that aligns generated stories with human storytelling preferences.

\vspace{-0.1cm}
\subsection{Human Evaluation}
Regarding the human assessment we follow the examples of numerous previous studies we measure the \textbf{Relevance to input images} or \textbf{Visual Grounding} and \textbf{Coherence} \cite{hu2020makesgoodstorydesigning,xu2021imagine, liu2023aog, wang2024scovistsocialinteractioncommonsense}, \textbf{Engagement and Interest} \cite{gervas2009computational, song2024contextawarevisualstorytellingvisual} and \textbf{Overall Likeness} \cite{chen2021commonsenseknowledgeawareconcept, wang2022rovistlearningrobustmetricsvisual}. Nonetheless, additionally to these criteria here we also assess the stories according to general \textbf{Language Quality and Style} criterion. 

We randomly choose 60 distinct visual story-sequences and the corresponding generated stories from 7 models (i.e., GLACNet, AREL, KG-Story, PR-VIST and MCSM+BART and two of our models, one with the T5 storyteller and one with BART\footnote{According to automatic metrics the frameworks that use Transformer as a captioner, both for T5 and BART perform better (see section \ref{sec:results} below), so we opt for this.}) to be evaluated through human assessment. 
The total number of participants were 10, which entails that each annotator evaluated 6 story-sequences, along with the machine generates storylines. 
Since the total number of criteria was 5 each participant should answer 30 questions throughout the questionnaire.
Lastly, the structure of this user-study was in the form of ranking, where each evaluator had to rank the 7 generated stories for each of the 5 examined criteria.

\subsection{LLM Evaluation}
Lately, with the advancement of \textit{Large Language Models (LLMs)} several studies have adopted this type of method for evaluating results and among those are also visual storytelling works \cite{ning2023albumstorytellingiterativestoryaware, chhun2024language}.  
In fact in many cases, LLMs, such as GPT-4 \cite{openai2024gpt4technicalreport}, have proven to provide comparable evaluation to those of humans \cite{, bai2023touchstoneevaluatingvisionlanguagemodels, liang2023largelanguagemodelsprovide}.
For these reasons, we also include LLM assessments in the present work. More specifically, we deploy GPT-4o and Claude 3.7 Sonnet \cite{wu2023comparativestudyopensourcelarge} for evaluating a sample of our machine generated stories. 
The evaluation criteria upon on which, these LLMs judged our stories, were the same as in human evaluation, meaning that the sample of image sequences was also $60$. 
Finally, the structure of the evaluation procedure remained the same as in human evaluation, where each LLM should rank the 7 produced storylines per examined criterion. The seven models tested were precisely the same as those mentioned for human evaluation.


\section{Results}
\label{sec:results}

\subsection{Automatic Metrics}
The performance, according to all the aforementioned automatic metrics, of the baselines methods and our frameworks on VIST test set is presented on Table \ref{tab:mainresults} (best scores are shown in bold and the second best are underlined).  
These results suggest that our frameworks, especially the version with Transformer+BART, are comparable or even better than many baselines on most automatic metrics. 
This includes even the powerful generalized multimodal LLM, StoryLLaVA, a model with more than $20$ times the size of our largest framework.

In particular, we can discern that in common automatic metrics that simply account for \textit{n-gram} overlap, like  BLEU-4 or METEOR, our model obtains worse score to most other baselines. 
Albeit, in more complicated metrics which take semantics or human-like criteria (such as coherence) under consideration, we can see that our method performs equally or even better than most state-of-the-art baselines.   
Especially for RoViST metrics which attempt to mimic human likeness (or judgment) our top method (Transformer+BART), poses a serious challenge to other models. 
This is clearly imprinted on the overall RoViST score, where our aforementioned framework is the only one that approaches the StoryLLaVA's performance ($81.9$ vs $82.9$)

\subsection{Human Judgment}

Regarding the human evaluation the results are provided on Table \ref{tab:human_eval_results}. These results are presented in the form of the average position (from $1^{st}$ till the $7^{th}$) that each model-contestant obtained, according to the evaluators assessments in their rankings.
This means that the lower the position that the stories of each method get, the better the quality of those stories are. 
From the foresaid table we can observe that our methods prevail against the other models in all criteria. 
In particular, Transf.+T5 is the best model regarding Visual Grounding, leaving second our Transf.+BART framework. Nonetheless, the latter is characterized as the top-1 model on all other criteria. 
Lastly, regarding the second best spot, this is obtained mainly by MCSM+BART approach (four out of five criteria), showcasing that BART has, in general, a very strong potential for cohesive visual storytelling. 

\begin{table*}[ht!]
    \centering
    \resizebox{\linewidth}{!}{ 
    \begin{tabular}{lcccc|ccc|cccc}
        \toprule
        \multirow{3}{*}{\Large{\textbf{Model}}} & \multicolumn{4}{c}{\large{\textbf{N-gram Overlap Metrics}}} & \multicolumn{3}{c}{\large{\textbf{Semantic \& Fluency Metrics}}} & \multicolumn{4}{c}{\large{\textbf{RoViST Metrics}}} \\
        & \textbf{BLEU-4} $\uparrow$ & \textbf{METEOR} $\uparrow$ & \textbf{ROUGE-L} $\uparrow$ & \textbf{CIDEr} $\uparrow$ 
        & \textbf{SPICE} $\uparrow$ & \textbf{BLEURT} $\uparrow$ & \textbf{PPL} $\downarrow$  
        & \textbf{VG} $\uparrow$ & \textbf{CHR} $\uparrow$ & \textbf{NR} $\uparrow$ & \makecell{\textbf{RoViST} $\uparrow$ \\ \textbf{((VG+CHR+NR)/3)}} \\
        \midrule
        
        AREL & \underline{13.8} & 35.2 & 29.9 & 9.5 & 9.0 & 32.6 & 12.0 & 67.2 & 57.9 & 83.4 & 69.5  \\
        GLACNet & 13.5 & 33.5 & 30.0 & 7.6 & 8.3 & 33.5 & 13.1 & 63.4 & 68.6 & 95.1 & 75.7  \\
        KG-Story & 10.0 & 31.5 & 25.2 & 9.8 & 7.2 & 32.3 & 46.1 & 60.8 & 65.2 & \textbf{99.9} & 75.3  \\
        ReCo-RL & 12.4 & 35.9 & 29.8 & 11.9 & 11.2 & 31.9 & 28.4 & 68.7 & 58.1 & 92.0 & 73.0  \\
        PR-VIST & 7.7 & 31.8 & 26.1 & 7.7 & 9.6 & 31.0 & 42.3 & 70.9 & 60.5 & 96.1 & 75.9  \\
        TAPM & 11.5 & \textbf{37.2} & \textbf{33.1} & \underline{13.8} & 10.0 & 33.4 & 18.3 & 71.5 & 67.1 & 90.5 &  76.4  \\
        MCSM+BART & 13.0 & \underline{36.1} & \underline{30.7} & 11.0 & 9.7 & 36.3 & 35.5 & 77.5 & 66.7 & 90.0 & 78.1  \\
        SCO-VIST & - & 27.5 & 22.1 & 6.2 & \underline{11.4} & 34.8 & 13.6 & 70.6 & \underline{75.9} & 92.2 & 79.6  \\
        Ca-VIST & \textbf{14.0} & 31.2 & 29.4 & 12.0 & 9.9 & 32.4 & \underline{10.5} & - & - & - & -  \\
        StoryLLaVA & - & 33.7 & 30.2 & \textbf{14.5} & - & - & 22.1\footnotemark & 75.1 & \textbf{83.3} & 90.4 & \textbf{82.9}  \\
        
        \midrule
        MLP+T5 & 2.4 & 17.0 & 16.3 & 11.3 & 9.8 & 31.6 & \textbf{10.1} & 72.1 & 55.1 & 86.4 & 71.2  \\
        Transformer+T5 & 2.6 & 18.2 & 17.4 & 10.8 & 10.2 & 32.2 & 10.7 & 73.3 & 58.9 & 87.4 & 73.2  \\
        MLP+BART & 6.9 & 22.5 & 19.0 & 11.9 & 10.5 & \underline{37.2} & 25.6 & \underline{77.6} & 64.7 & 96.1 & 79.5  \\
        Transformer+BART & 13.3 & 27.6 & 23.8 & 12.6 & \textbf{11.6} & \textbf{38.9} & 28.4 & \textbf{78.0} & 70.8 & \underline{96.7} & \underline{81.9}  \\
        \bottomrule
    \end{tabular}}
    \vspace{-0.2cm}
    \caption{\centering \small {Performance on automatic metrics of the baseline models (above the horizontal line) and our four frameworks (below the horizontal line). 
    The vertical lines denote metrics that measure similar traits. 
    The first 4 metrics measure n-gram overlap, the following 3 measure semantics and fluency and RoViST metrics capture human-likeness. 
    PPL: Perplexity; VG: Visual Grounding; CHR: Coherence; NR: non-Redundancy. The best scores are in \textbf{bold} and the second best are \underline{underlined}.}}
    \label{tab:mainresults}
\end{table*}
\footnotetext{This result is based on a very small sample from the model's generated stories.}

\begin{table*}[ht!]
    \centering
    \hspace*{-0.8cm} 
    \resizebox{1.08\textwidth}{!}{ 
    \begin{tabular}{lcc|ccc|cccc|ccccc|cc}
        \toprule
        \Large{\textbf{Model}} & \makecell{\textbf{Avg. Story} \\ \textbf{Length} $\uparrow$ } & \makecell{\textbf{Avg. Sent.} \\ \textbf{Length} $\uparrow$} & \Large{$\frac{|\mathbf{V}|}{n} \uparrow $} & \Large{$\frac{\mathbf{N_{tok}}}{n} \uparrow$} & \textbf{TTR} $\uparrow$ & \makecell{$(\%)$ \\ \textbf{Nouns}} & \makecell{$(\%)$ \\ \textbf{Verbs}} & \makecell{  $(\%)$ \textbf{Pro-} \\ \textbf{Nouns}} & \makecell{$(\%)$ \\ \textbf{ADJ}} & $\mathbf{rep_1}$ $\downarrow$ & $\mathbf{rep_2}$ $\downarrow$ & $\mathbf{rep_3}$ $\downarrow$ & $\mathbf{rep_4}$ $\downarrow$ & \textbf{Diversity} $\uparrow$ & $\mathbf{K}$ $\downarrow$ & $\mathbf{H}$ $\uparrow$ \\
        \midrule

        Humans & 58.9 & 11.11 & \underline{3.257} & 57.24 & \underline{5.69} & 19.78 & 12.65 & 10.17 & 6.55 & \underline{19.44} & 2.62 & 0.71 & 0.27 & 96.42 & \textbf{126.09} & \underline{5.11} \\
        \midrule
        AREL & 44.8 & 8.84 & 0.423 & 45.75 & 0.92 & 21.32 & 9.08 & 8.74 & 7.39 & 46.44 & 24.02 & 17.00 & 12.51 & 55.16 & 422.32 & 4.25 \\
        GLACNet & 35.2 & 7.02 & 0.271 & 35.12 & 0.77 & 17.68 & 8.80 & 12.09 & 10.31 & 29.44 & 5.52 & 2.13 & 0.70 & 91.81 & 315.21 & 4.38 \\
        KG-Story & 32.3 & 6.53 & 0.196 & 32.47 & 0.60 & 15.61 & 13.45 & 15.72 & 8.55 & \textbf{12.90} & \textbf{0.03} & \textbf{0.002} & 0.0 & \textbf{99.96} & 197.59 & 4.64 \\
        ReCo-RL & 49.3 & - & - & - & - & - & - & - & - & 33.58 & 3.14 & 0.11 & 0.02 & 97.27 & - & - \\
        PR-VIST & 52.3 & 9.72 & 0.369 & 52.29 & 0.71 & 18.91 & 9.46 & 11.71 & 7.17 & 29.50 & 5.39 & 1.27 & 0.41 & 93.01 & 215.14 & 4.91 \\
        TAPM & 51.2 & - & - & - & - & - & - & - & - & 36.16 & 10.02 & 5.16 & 2.89 & 82.87 & - & - \\
        MCSM+BART & 56.7 &\underline{11.61} & \textcolor{red}{2.916} & \underline{64.35} & \textcolor{red}{4.53} & 20.73 & 12.10 & \textcolor{red}{9.94} & 6.35 & 36.63 & 8.84 & 2.88 & 1.13 & 87.53 & 249.19 & 4.95 \\
        Ca-VIST & - & - & - & - & - & - & - & - & - & 24.40 & 2.04 & 0.23 & \textcolor{red}{0.05} & 97.69 & - & - \\
        StoryLLaVA\footnotemark & \textbf{160.7} & \textbf{17.21} & \textbf{39.13} & \textbf{94.33} & \textbf{41.48} & 23.89 & 11.73 & 6.50 & 8.69 & 27.98 & \textcolor{red}{2.51} & \textcolor{red}{0.34} & 0.0 & \textcolor{red}{97.17} & \underline{\textcolor{red}{132.89}} & \textbf{5.31} \\

        \midrule
        MLP+T5 & 45.3 & 8.80 & 0.488 & 44.60 & 1.09 & 23.45 & 10.64 & 10.07 & 4.78 & 46.64 & 31.66 & 25.03 & 19.78 & 41.09 & 517.12 & 4.04\\
        Transf.+T5 & 47.6 & 8.81 & 0.606 & 43.80 & 1.38 & 23.66 & 10.73 & 9.79 & 4.89 & 44.90 & 29.60 & 22.99 & 17.86 & 44.52 & 479.07 & 4.11 \\
        MLP+BART & 55.1 & 10.20 & 1.100 & 51.16 & 2.15 & 21.18 & 12.27 & 9.25 & 6.27 & 23.26 & 2.35 & \underline{0.08} & 0.0 & 97.57 & 159.74 & 5.01\\
        Transf.+BART & \underline{\textcolor{red}{60.2}} & \textcolor{red}{10.88} & 2.016 &  \textcolor{red}{55.46}  & 3.64 & \textcolor{red}{19.93} & \textcolor{red}{12.33} & 9.64 & \textcolor{red}{6.61} & \textcolor{red}{20.52} & \underline{1.83} & 0.09 & 0.0 & \underline{98.08} & 139.24 & \textcolor{red}{5.05}\\
        \bottomrule
    \end{tabular}}
    \vspace{-0.2cm}
    \caption{\centering \small {Performance on lexical metrics of all baseline models (above the horizontal line) and our four frameworks (below the horizontal line). 
    The vertical lines denote metrics that measure similar traits. 
    The first 2 metrics measure length of generated text, the following 3 are token-based metrics. Then, we have POS measuring metrics, followed by n-gram repetition metrics which account for lexical diversity and lastly we have measures for linguistic richness and unpredictability. $|V|$: Number of unique tokens (1-grams); $N_{tok}$: Total number of tokens generated; $n$: number of stories examined per model; TTR: Type-Token Ratio; ADJ: Adjectives; $rep_n$: repetition rate of n-grams; $K$: Yule’s measure for lexical richness; $H$: Shannon's entropy for lexical unpredictability.
    The best scores are in \textbf{bold} and the second best are \underline{underlined}.
    The \textcolor{red}{red highlighted} scores denote the model that minimized the difference between its own score and the respective human score on that specific metric.}}
    \label{tab:lexical_results}
    \vspace{-0.3cm}
\end{table*}
\footnotetext{The results of StoryLLaVA in this table arose from a very small sample from its own generated stories and its lexical scores are majorly affected (eg. $\frac{|V|}{n}$=$39.13$ or TTR = $41.48$).}

\subsection{LLM Assessment}
The aggregated results from LLM evaluation, are presented on Table \ref{tab:llm_eval}. Since each LLM ranked on a 1 to 7 scale the models that generated stories, the results pertain to the average position that each model acquired according to each LLM's judge. 
Consequently, the lower the obtained average position by each model is, the better its produced stories according to that LLM over its $60$ assessments are.
From these results, we can discern a similar grading pattern to that from human evaluation. More specifically, both LLMs rank our two contestant methods consistently in lower positions than the rest of the methods. 
Our Transf.+T5 model obtains the lowest spot in Visual Grounding (like in the human evaluation), while in all other criteria, Transf.+BART has the advantage. 
In addition, Claude 3.7 ratings seem to consistently favor more the stories of our Transf.+BART model across all criteria, by giving lower ranks than GPT-4o.  
Finally, regarding the other models, MCSM+BART is the only one that is ranked among the top-2 methods, especially when we focus on Claude's 3.7 assessment.

\begin{table}[ht]
    \centering
    \small
    \hspace*{-0.2cm} 
    \begin{tabular}{lccccc}
        \toprule
        \toprule
        \textbf{Methods} & \textbf{VG ↓} & \textbf{CF ↓} & \textbf{EI ↓} & \textbf{LQS ↓} & \textbf{OL ↓} \\
        \midrule
        AREL        & 4.62  & 5.04  & 5.23  & 4.92  & 4.88  \\
        GLACNet     & 4.45  & 3.77  & 4.67  & 4.75  & 4.62  \\
        KG-Story    & 5.06  & 4.83  & 4.90  & 4.81  & 4.81  \\
        PR-VIST     & 4.58  & 4.67  & 4.60  & 4.44  & 4.48  \\
        MCSM+BART   & 3.52  & \underline{3.30}  & \underline{2.98} & \underline{2.94}  & \underline{3.19} \\
        \midrule
        Transf.+T5  & \textbf{2.79} & 3.33 & 3.23  & 3.81 & 3.62  \\
        Transf.+BART & \underline{2.98} & \textbf{3.06} & \textbf{2.39} & \textbf{2.33}  & \textbf{2.37}  \\
        \bottomrule
        \bottomrule
    \end{tabular}
    \vspace{-0.2cm}
    \caption{\centering \small{Human evaluation on the quality of stories generated by five state-of-the-art baselines (above the
    horizontal line) and by our best models with T5 \& BART storytellers (below the horizontal line). Best scores are in bold and the second best are underlined. VG: Visual Grounding, CF: Coherence \& Flow, EI: Engagement \& Interest, LQS: Language Quality \& Style, OL: Overall Likeness.}}
    \label{tab:human_eval_results}
\end{table}

\begin{table*}
\centering
\resizebox{\linewidth}{!}{ 
\begin{tabular}{lcccccccccc}
\toprule
\toprule
\multirow{2}{*}{\textbf{Methods}} & \multicolumn{2}{c}{\textbf{Visual Grounding} $\downarrow$}  & \multicolumn{2}{c}{\textbf{Coherence} $\downarrow$} & \multicolumn{2}{c}{\textbf{Engagement} $\downarrow$} & \multicolumn{2}{c}{\textbf{Language} $\downarrow$} &  \multicolumn{2}{c}{\textbf{Overall} $\downarrow$} \\ [0.5ex]
& GPT-4o & Claude 3.7 & GPT-4o & Claude 3.7 & GPT-4o & Claude 3.7 & GPT-4o & Claude 3.7 & GPT-4o & Claude 3.7 \\ [0.5ex]
\hline
AREL & 4.35 & 4.66 & 4.72 & 4.40 & 4.80 & 4.97 & 4.69 & 4.94  & 4.64 & 4.85 \\ [0.3ex]
GLACNet & 4.72 & 4.51 & 4.96 & 4.74 & 5.26 & 5.03 & 5.21 & 4.86 & 5.30 & 4.92 \\[0.5ex]
KG-Story & 4.78 & 5.76 & 4.89 & 5.88 & 4.85 & 5.94 & 4.76 & 5.95 & 4.62 & 5.80 \\ [0.3ex]
PR-VIST & 5.66 & 4.83 & 5.58 & 4.71 & 5.51 & 4.12 & 5.40 & 4.29 & 5.53 & 4.53 \\ [0.3ex]
MCSM+BART & 3.08 & 2.98 & 2.89 & \underline{2.90} & \underline{2.73} & \underline{2.72} & 3.11 & \underline{2.77} & 2.95 & \underline{2.81} \\ [0.3ex]
\hline
Transf.+T5 & \textbf{2.65} & \textbf{2.61} & \underline{2.79} & 3.29 & 2.76 & 3.49 & \underline{2.73} & 3.48 & \underline{2.80} & 3.33 \\ [0.3ex]
Transf.+BART & \underline{2.76} & \underline{2.65} & \textbf{2.50} & \textbf{1.90} & \textbf{2.09} & \textbf{1.74} & \textbf{2.10} & \textbf{1.71} & \textbf{2.16} & \textbf{1.75} \\
\bottomrule
\bottomrule
\end{tabular}}
\vspace{-0.2cm}
\caption{\centering \small {LLM evaluation from GPT-4o \& Claude 3.7 on the story quality from 5 state-of-the-art baselines (above the horizontal line) and stories generated from our best frameworks with T5 \& BART storytellers (below the horizontal line). The best scores are in bold and the second best are underlined.}}
\label{tab:llm_eval}
\end{table*}


\section{Analysis}
\label{sec:analysis}

\subsection{Linguistic Analysis of stories}
\label{subsec:results_mapping}
Besides evaluating our models on automatic metrics we carried out an extensive analysis regarding the lexical characteristics of our generated stories. 
In contrast to Table \ref{tab:mainresults}, for this experiment we keep only models for which we have directly access to their stories or their authors have conducted similar lexical analysis. 
For this reason, we exclude SCO-VIST model. However, here we include the original VIST test stories written by humans and we use them as a central baseline. 

Initially, we measure the \textit{average story length} and \textit{average sentence length} for each model. Then we proceed with token-based analysis by measuring the \textit{vocabulary size}, $|V|$, of each models (total number of unique uni-grams produced) and the \textit{total number of tokens} produced $N_{tok}$. 
We normalize this numbers with the total numbers of available stories per model, $n$\footnote{This number is different for methods from other works. For example, our test sample was 3,338, while for AREL we generated 2,017 stories.}. Finally, we compute the \textit{token-type ratio} (TTR) as: $\text{TTR} = \frac{|V|}{N_{tok}}$.
 
Furthermore, we investigate the characteristics of the generated stories in terms of part-of-speech (POS) tags, which can provide useful insights on the structure of the machine generated language.
These are \textit{Nouns}, \textit{Verbs}, \textit{Pronouns} and \textit{Adjectives} (\textit{ADJ}). 
Additionally, we assess text degeneration and linguistic diversity according to \cite{su2022contrastiveframeworkneuraltext}, where firstly, we measure story-level repetition by computing the duplicate n-grams as: $rep_n = 100 \times (1 - \frac{|\text{unique $n$-grams}|}{|\text{total $n$-grams}|})$, and then we calculate token \textit{diversity} as: 

\begin{equation}
diversity = \prod_{n=2}^4  \left( 1-\frac{rep_n}{100} \right).
\end{equation}

Finally, we measure two other diversity metrics, namely \textit{Yule's} $K$ \cite{tanaka-ishii-aihara-2015-computational}, which accounts for lexical richness and \textit{Shannon's entropy} ($H$) \cite{shannon1948mathematical}, which quantifies the unpredictability of word choice in the story-text.
The formulas for these two metrics, are the following:
\vspace{-0.1cm}
\begin{subequations}
\begin{align}
K = 10^4 \times \frac{\sum f_i^2 - L}{L^2}  \\ 
H = - \sum_{i=1}^{\Tilde{L}} p_i \log_2(p_i),
\end{align}
\vspace{-0.1cm}
\end{subequations}
where $L$ is the number of words in each story, $f_i$ is the frequency of the $i^{th}$ word of a story, $p_i$ is the probability of the $i^{th}$ word occurring in the story and $\Tilde{L}$ is the number of unique words within a story.
The accumulated results regarding these lexical traits, are presented on Table \ref{tab:lexical_results}.
From this table we can see vividly that in terms of lexicology StoryLLaVA dominates most metrics, by achieving the highest (lowest) scores, whilst in terms of word-repeatability and diversity, KG-Story is the one that comes first.
Nevertheless, a model that generates the most lengthy story or the one with lowest word-repeatability does not guarantee us, that it also gives the most overall completed story nor the one that is more human alike.

\subsection{The notion of Ideality}
\label{subsec:ideality} 
On the top of Table \ref{tab:lexical_results}, we report the performance of ground-truth stories, which are written by humans (from VIST dataset) in all these lexical metrics. 
We then, make the convention that these ground-truth stories represent the \say{ideal} kind of storytelling and to that end, we utilize their results in order quantify how far are the machine generated stories from human-like written text. 
For an individual metric, we highlight in red, the model's score that minimizes the absolute distance from the respective human score. We can formulate this \textit{closer-to-humans score}($ch$), by using the following equation: 
\begin{equation}
    ch = \min_{ms \in \mathbb{M}} |ms-hs|
    \label{eq:closer_to_humans}
    \vspace{-0.2cm}
\end{equation}
In equation (\ref{eq:closer_to_humans}), we denote with $hs$ the score of human written stories on any lexical metric, with $ms$ the respective scores of the machine generated stories and with $\mathbb{M}$ the total set of models that are tested (all models on Table \ref{tab:lexical_results}).
Likewise, we can keep track of how close to humans scores are the individual model's scores (row scores).
This novel metric is called \say{ideality} and indicates how far a generated story is from the respective human one, on the linguistic traits that are presented on Table \ref{tab:lexical_results}.
Formally, we can expound a model's \textit{ideality} as: 
\vspace{-0.1cm}
\begin{equation}
    ideality = \sum_{ms, hs \in \mathbb{S}} \frac{1}{\sigma(|ms-hs|)},
    \label{eq:ideality}
\vspace{-0.1cm}
\end{equation}
where $\mathbb{S}$ is the set of metrics displayed on Table \ref{tab:lexical_results} and $\sigma$ is the \textit{sigmoid} function. In mathematical terms, we can think of \textit{ideality} as a function of scores that a model gets on a set of metrics. 
We can define this function as follows: $I(ms): \mathbb{R} \rightarrow (|\mathbb{S}|,2|\mathbb{S}|]$, where $ |\mathbb{S}|$ represents the number of elements in $\mathbb{S}$. 
For our case, we have $16$ evaluation criteria from Table \ref{tab:lexical_results}, so $|\mathbb{S}|=16$, and therefore $I$ can get values in the interval $(16,32]$\footnote{The domain value of $I$ is between $|\mathbb{S}|$ and $2|\mathbb{S}|$ by the definition of the function at eq.\ref{eq:ideality}. 
In the worst case, all of the model's results will be $0$ (practically impossible), so the \textit{sigmoid} will take only the human indicators as inputs, and thus we get: $ \sum_{|\mathbb{S}|} \frac{1}{\sigma(|hs|)} = |\mathbb{S}| \times \frac{1}{\sigma(|hs|)}. \approx |\mathbb{S}|$, if we always have $|hs|>>1$ for all human scores. 
Conversely, if we have a perfect model that succeeds exactly at human level (also impossible), then the inputs of \textit{sigmoid} will be $|hs-hs|=0$ and thus we get: $\sum_{|\mathbb{S}|} \frac{1}{\sigma(0)}= |\mathbb{S}| \times\frac{1}{0.5} = 2|\mathbb{S}|$.}

However, in many cases not all examined metrics are of equal importance. For example, $\text{rep}_1$ does not give us the same information as $\text{rep}_4$, let alone \textit{diversity}. 
As a consequence, we propose the improvement of \textit{weighted ideality}, which assigns to each of the examined metrics different weight: 
\begin{equation}
    W_{ideality} = \sum_{ms, hs \in \mathbb{S},\\ i = 1}^{|\mathbb{S}|} \frac{w_i}{\sigma(|ms-hs|)}.
    \label{eq:W_ideality}
\end{equation}
Depending on how much weight we want our more important metrics to gain, we set a restriction for the weight vector $\mathbf{W} = [w_1, \ldots, w_{|\mathbb{S}|}]$ as follows: 
\vspace{-0.2cm}
\begin{itemize}
    \item $||\mathbf{W}||_1 = |\mathbb{S}|$ for lower variability, where each metric gets a weight around $1$.
    \vspace{-0.2cm}
    \item $||\mathbf{W}||_2 = |\mathbb{S}|$ for higher variability, where each metric gets a weight around $\sqrt{|\mathbb{S}|}=4$.
\end{itemize}

\subsubsection{Results from Ideality}
\begin{figure}[H]
    \centering
    \begin{subfigure}[b]{0.99\linewidth}
        \includegraphics[width=0.99\linewidth]{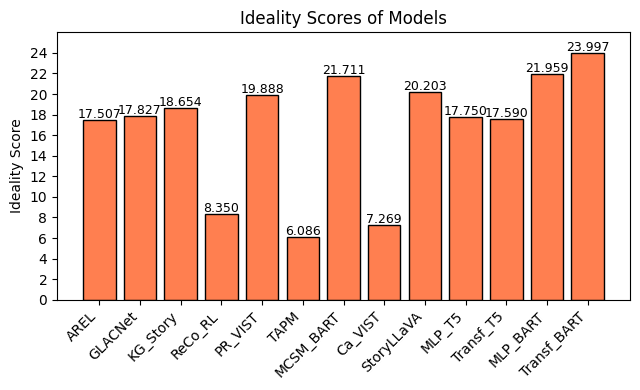}
    \end{subfigure}
    \begin{subfigure}[b]{0.99\linewidth}
        \includegraphics[width=0.99\linewidth]{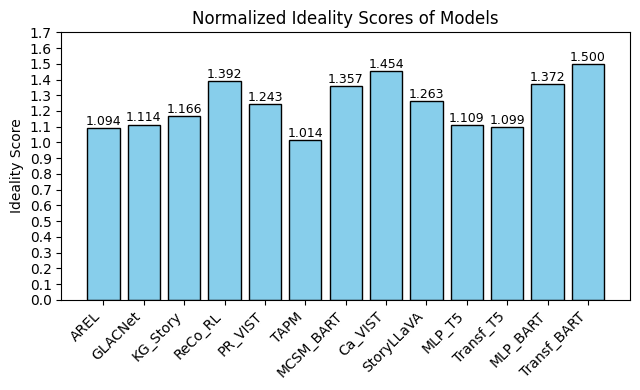}
    \end{subfigure}
    \vspace{-0.8cm}
    \caption{ \centering Ideality results for all models appearing on Table \ref{tab:lexical_results}. Top: Raw ideality scores per model. Bottom: Ideality scores per model normalized over the number of metrics that actually each model was scored.}
    \label{fig:idealities}
    \vspace{-0.3cm}
\end{figure}

Looking back on Table \ref{tab:lexical_results} we can see that out our model with Transformer and BART, gets 8 out of the 16 measured metrics, when we focus on the $ch$ score. 
Following this, we have the StoryLLaVA which gets 4 times the $ch$ metric, MCSM+BART that gets 3 and Ca-VIST with 1. 
To further assess the human-likeness of our stories, we illustrate on Fig. \ref{fig:idealities} the (unweighted) ideality results from all models that were evaluated on Table \ref{tab:lexical_results}. 
In the top of that figure we have the non-normalized ideality scores per model, which are not representative if all models are not evaluated on precisely the same number of metrics\footnote{this is valid for Table \ref{tab:lexical_results}, as some models have be evaluated to sixteen metrics while others only on five.}. 
For this reason, we also depict the normalized ideality scores, which essentially consists of the ideality score divided by the number of metrics on which a model was assessed.  

\vspace{-0.2cm}

\begin{figure*}[ht!]
    \centering
    \begin{subfigure}[b]{0.94\linewidth}
        \hspace{0.4cm}\includegraphics[width=0.94\linewidth]{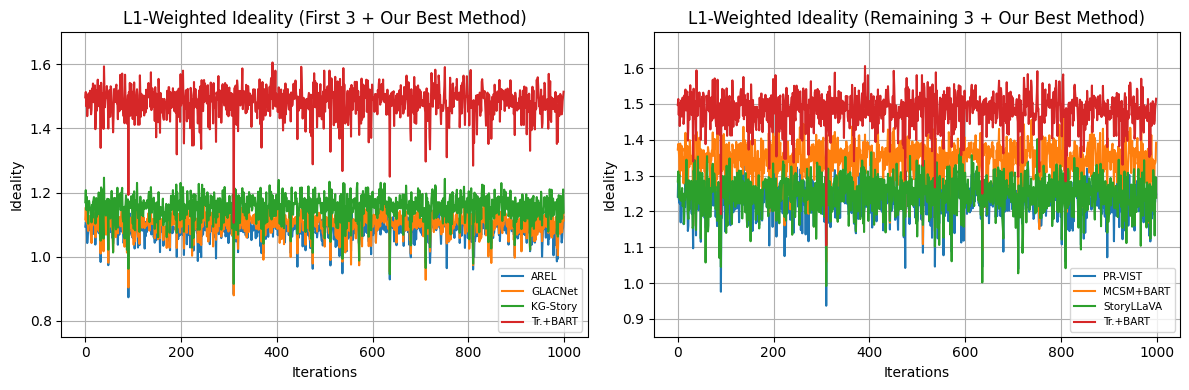}
    \end{subfigure}
    \begin{subfigure}[b]{0.94\linewidth}
        \hspace{0.4cm}\includegraphics[width=0.94\linewidth]{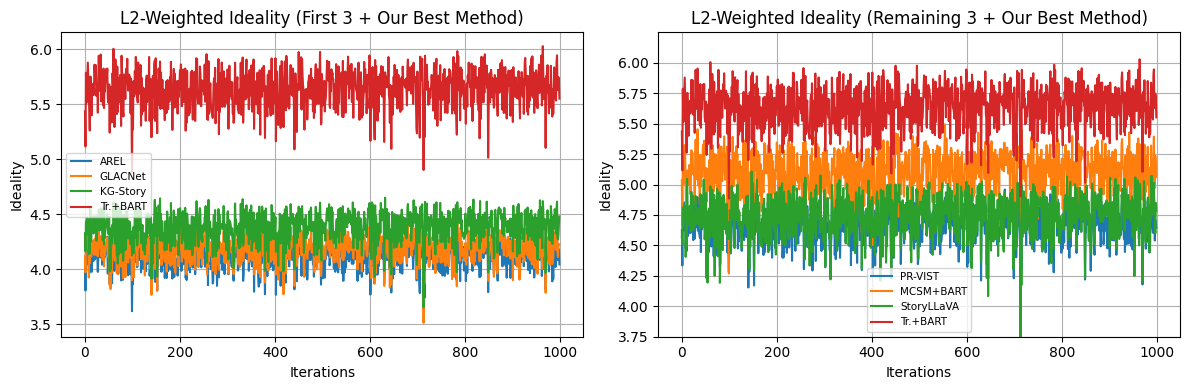}
    \end{subfigure}
    \vspace{-0.4cm}
    \caption{ \centering Weighted ideality results, over 1000 random runs, from our best performing method against 6 models that were evaluated on all 16 metrics appearing on Table \ref{tab:lexical_results}. Top: L1-weighted ideality. Bottom: L2-weighted ideality.}
    \label{fig:weighted_idealities}
\end{figure*}

The results from Fig. \ref{fig:idealities} indicate that our model with Transformer mapping network as captioner and the BART storyteller, is not only superior to all other alternative of our frameworks, but also from all other baselines in terms of ideality. 
This stands even for models that were evaluated on fewer metrics than ours (ie. Ca-VIST), where it is easier to keep $ch$ on a lower level.  
For weighted ideality, instead of arbitrarily choosing which metric from Table \ref{tab:lexical_results} is more important, we opt to run an experiment of $1000$ runs, where we let the vector $\mathbf{W}$ to be randomly set under the restrictions of L1 and L2-norms.
As as consequence, each metric from Table \ref{tab:lexical_results} gets random weights on an iterative procedure, where for each run we measure the corresponding ideality score of every model. 
The final results are depicted on Fig. \ref{fig:weighted_idealities}, where we compared the weighted ideality scores of our top performing framework (Transf.+BART) against all models that were evaluated on the same number of metrics (ie. AREL, GLACNet, KG-Story, PR-VIST, MCSM+BART and StoryLLaVA). 
In the top row of that figure we can see the results from  L1-$W_{ideality}$, while in the bottom row we can see the results from the L2-$W_{ideality}$. 
In both occasions, it is clear that our Transf.+BART framework outperforms the remaining baselines, having on average a pure heading margin across the 1000 random runs.

\subsection{Qualitative Analysis of Story-Examples}
\begin{figure*}[ht!]
    \centering
    \vspace{-0.5cm}
    \hspace{-1cm}\footnotesize
    \setlength{\tabcolsep}{2pt} 
    \renewcommand{\arraystretch}{1.12} 
    \begin{tabular}{>{\centering\arraybackslash}m{1.2in}>
    {\centering\arraybackslash}m{1in} >{\centering\arraybackslash}m{1in} >{\centering\arraybackslash}m{1in} >{\centering\arraybackslash}m{1in} >{\centering\arraybackslash}m{1in}}
    
        \small \textsf{\textbf{Visual Story}:} &
        \includegraphics[width=1in]{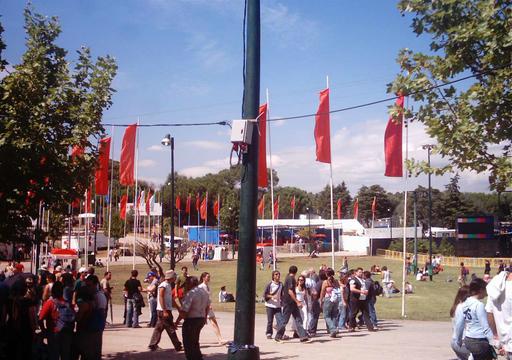} & 
        \includegraphics[width=1in]{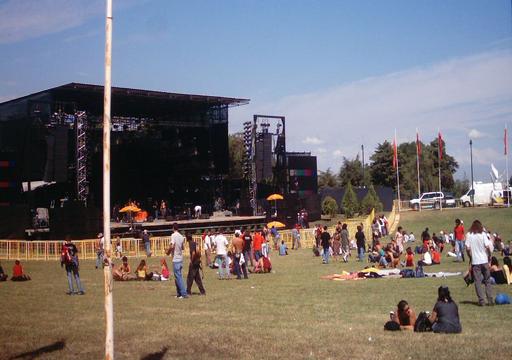} & 
        \includegraphics[width=1in]{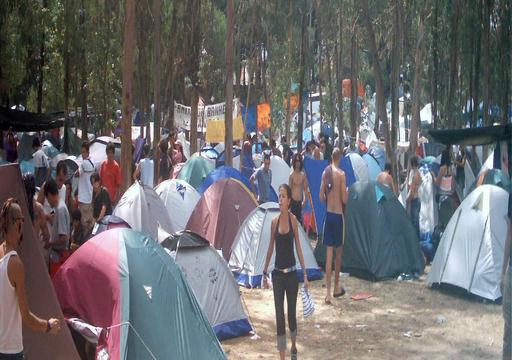} & 
        \includegraphics[width=1in]{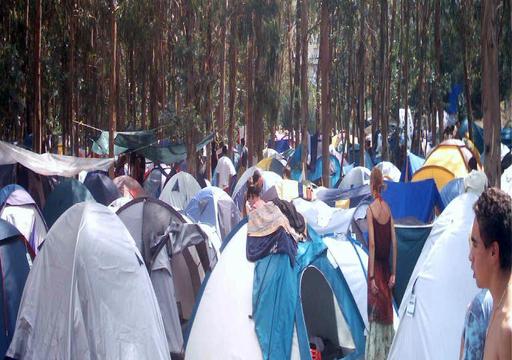} & 
        \includegraphics[width=1in]{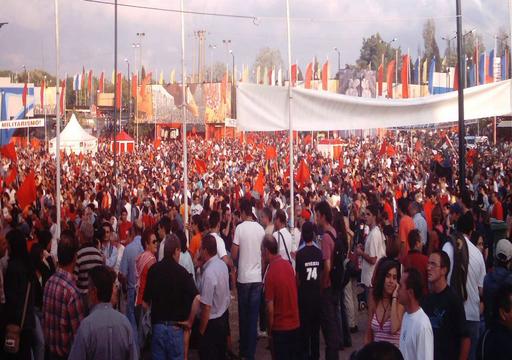} \\
        
        \small \textbf{AREL}: &
        \scriptsize I went to the \textcolor{red}{parade} yesterday. & 
        \scriptsize \hl{There were a lot of people there}. & 
        \scriptsize \hl{There was a lot of people there}. & 
        \scriptsize \hl{There were a lot of people there}. & 
        \scriptsize It was a great day for a group of \hl{people}. \\
        
        \small \textbf{GLACNet}: &
        \scriptsize The \textcolor{red}{parade} was a lot of fun. & 
        \scriptsize There were many people there. & 
        \scriptsize Everyone was very excited. & 
        \scriptsize It was \hl{a great time}. & 
        \scriptsize I had \hl{a great time}. \\
        
        \small \textbf{KG-Story}: &
        \scriptsize I went to the \textcolor{red}{city} today. & 
        \scriptsize Everyone was having a great time. & 
        \scriptsize There were many people gathered. & 
        \scriptsize They all watched \textcolor{blue}{as well}. & 
        \scriptsize \textcolor{blue}{Afterwards}, we had food. \\
        
        \small \textbf{PR-VIST}: &
        \scriptsize It was very \textcolor{red}{cold} outside. & 
        \scriptsize There were so many people in the \textcolor{red}{town}. & 
        \scriptsize They broke up with other people today. & 
        \scriptsize It was a long day. & 
        \scriptsize The \textcolor{red}{beach} was full of \textcolor{red}{umbrellas}. \\
        
        \small \textbf{MCSM+BART}: &
        \scriptsize I attended a special day for our city and the crowd was large. & 
        \scriptsize The band was playing and the crowd was set up. & 
        \scriptsize One of the many tents we had set up to watch the event. & 
        \scriptsize Some people were in tents that looked as though they were going to be challenged. & 
        \scriptsize I had to say hello to one of the men in a tent and as I approached the crowd, I could see that I had lost my seat. \\
        
        \small \textbf{Transformer+T5 (Ours)}: &
        \scriptsize  I went to the festival last week.  & 
        \scriptsize \hl{There were a lot of people there}.  & 
        \scriptsize \hl{I had a great time there}. & 
        \scriptsize \hl{There were} also \hl{a lot of people there}. & 
        \scriptsize And \hl{I had a great time there}.\\
        
        \small \textbf{Transformer+BART (Ours)}: &
        \scriptsize A huge crowd built up in anticipation for this band. & 
        \scriptsize The concert spilled out into the streets, and the sound was deafening as they set up their instruments. & 
        \scriptsize There were so many people set up, that they were spread out all over the street. & 
        \scriptsize They were so lucky to be the first ones to arrive; they had the best spot. & 
        \scriptsize Everyone was singing and dancing, and it was hard to contain themselves. \\
        
        \small \textsf{\textbf{Visual Story}:} &
        \includegraphics[width=1in]{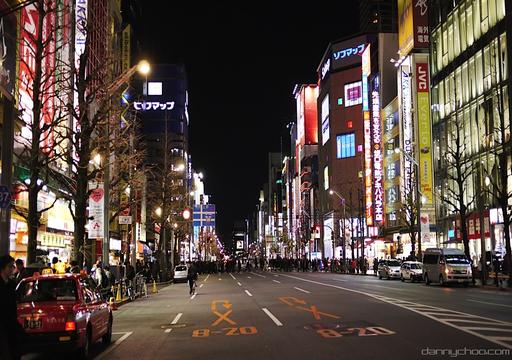} & 
        \includegraphics[width=1in]{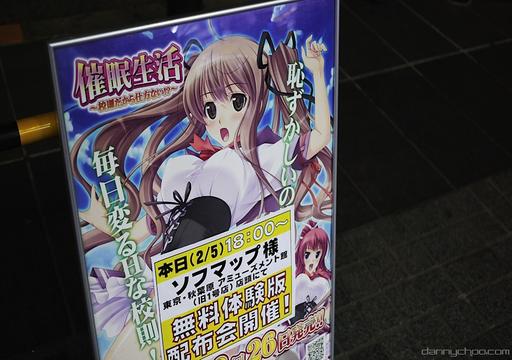} & 
        \includegraphics[width=1in]{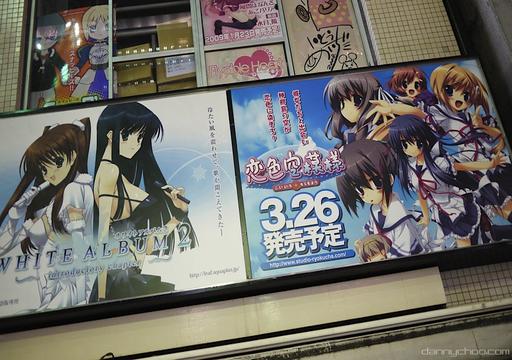} & 
        \includegraphics[width=1in]{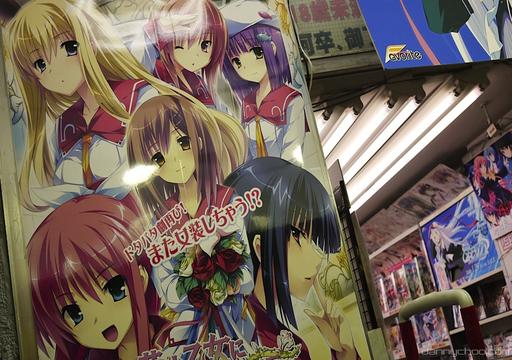} & 
        \includegraphics[width=1in]{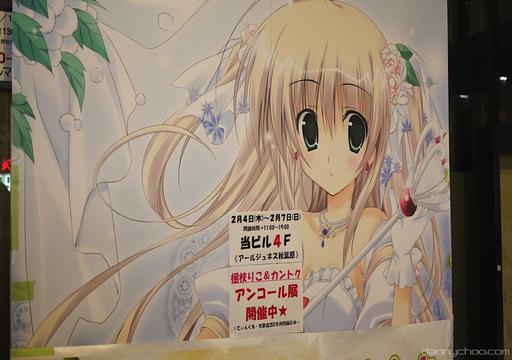} \\
        
        \small \textbf{AREL}: &
        \scriptsize I went to the city last week. & 
        \scriptsize There were many interesting things to see. & 
        \scriptsize \hl{There was a lot of} interesting \hl{art} on the wall. & 
        \scriptsize \hl{There was} \textcolor{blue}{also} \hl{a lot of art}. & 
        \scriptsize This is a picture of a \textcolor{red}{man}. \\
        
        \small \textbf{GLACNet}: &
        \scriptsize The city was a great place to visit. & 
        \scriptsize There were many interesting signs. & 
        \scriptsize \textcolor{red}{People} had fun and exhibits. & 
        \scriptsize They even had a museum. & 
        \scriptsize This was a very cool art piece. \\
        
        \small \textbf{KG-Story}: &
        \scriptsize I had a great time in the city. & 
        \scriptsize There were many different artworks. & 
        \scriptsize This one was my favorite. & 
        \scriptsize It \textcolor{blue}{also} made me very happy. & 
        \scriptsize \textcolor{blue}{Afterwards}, we all \textcolor{red}{hung out}. \\
        
        \small \textbf{PR-VIST}: &
        \scriptsize Our vacation in town was a great place to visit. & 
        \scriptsize The kids were so excited for their first game. & 
        \scriptsize They played the game on television and enjoyed it. & 
        \scriptsize They saw some souvenirs and art in the store. & 
        \scriptsize I went to the eye store and it was a great time with the paintings. \\
        
        \small \textbf{MCSM+BART}: &
        \scriptsize The location is a beautiful city and quite unique. & 
        \scriptsize It has wonderful things to do, like this downtown art exhibit. & 
        \scriptsize The interiors of the art shows a famous place like McDonald's. & 
        \scriptsize It is filled with beautiful artwork as well. & 
        \scriptsize People even come and visit the museum to see the weird things. \\
        
        \small \textbf{Transformer+T5 (Ours)}: &
        \scriptsize I went to the \textcolor{black}{mall} to do some shopping. & 
        \scriptsize They had a lot of anime there. & 
        \scriptsize I bought some of them.  & 
        \scriptsize I \textcolor{blue}{also} bought some manga. & 
        \scriptsize It was a very fun day. \\
        
        \small \textbf{Transformer+BART (Ours)}: &
        \scriptsize I went to Akihabara last night and had a great time; the streets were crowded with people, which did not surprise me. & 
        \scriptsize I found my Wifu, \textcolor{blue}{though}, which I have to admit was pretty awesome. & 
        \scriptsize I can't wait to see the next one. & 
        \scriptsize I browsed a bit more and saw that there were also action figures available. & 
        \scriptsize \textcolor{blue}{Of course}, I grabbed a copy of "Big Boobed Himoko Chan," which I've always wanted to obtain.  \\

        \small \textsf{\textbf{Visual Story}:} &
        \includegraphics[width=1in]{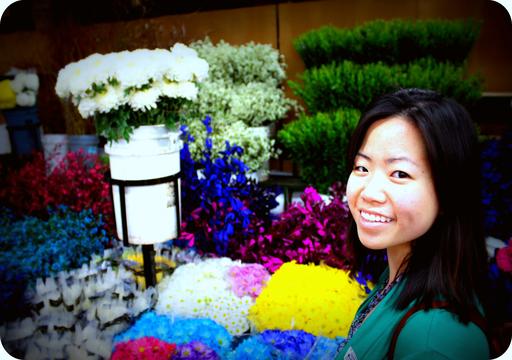} & 
        \includegraphics[width=1in]{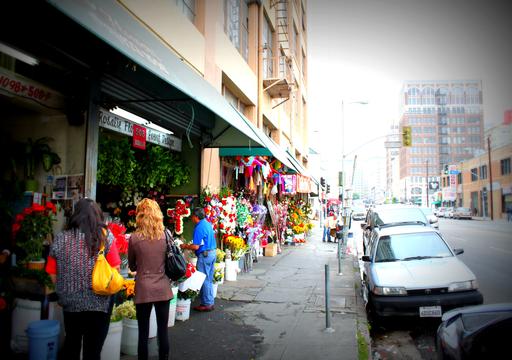} & 
        \includegraphics[width=1in]{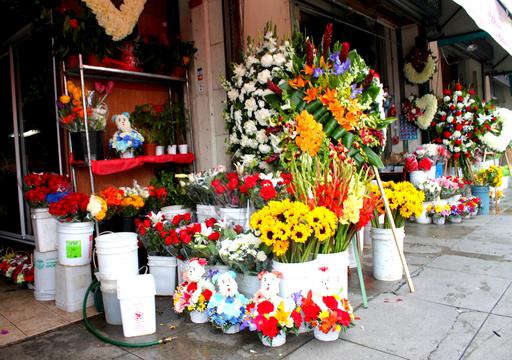} & 
        \includegraphics[width=1in]{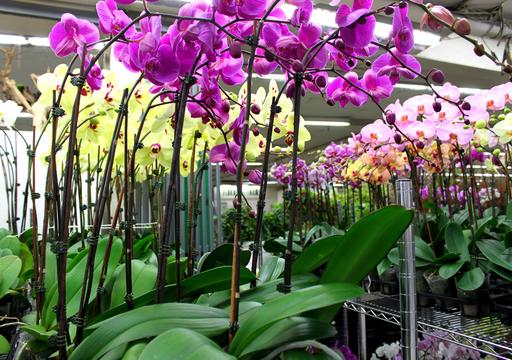} & 
        \includegraphics[width=1in]{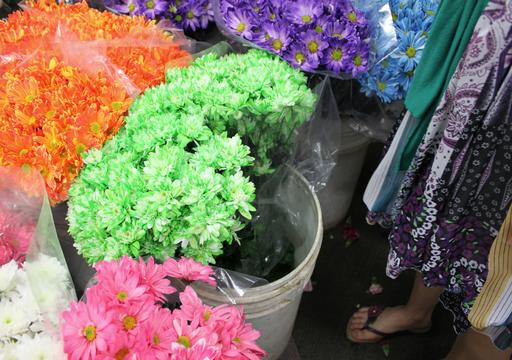} \\
        
        \small \textbf{AREL}: &
        \scriptsize I went to the market today. & 
        \scriptsize The \hl{flowers and flowers} are very beautiful. & 
        \scriptsize \hl{I had a great time there}. & 
        \scriptsize The \hl{flowers} were decorated with \hl{flowers}. & 
        \scriptsize \hl{I had a great time there}. \\
        
        \small \textbf{GLACNet}: &
        \scriptsize The \textcolor{red}{town} was filled with \textcolor{red}{people}. & 
        \scriptsize There were \hl{flowers and flowers}. & 
        \scriptsize They had a variety of fresh \textcolor{red}{fruits}. & 
        \scriptsize Some were very colorful. & 
        \scriptsize Others were selling. \\
        
        \small \textbf{KG-Story}: &
        \scriptsize We went to the botanical garden today. & 
        \scriptsize There were many beautiful flowers. & 
        \scriptsize It was very pretty. & 
        \scriptsize I had a great time. & 
        \scriptsize Some of my favorite plants. \\
        
        \small \textbf{PR-VIST}: &
        \scriptsize I bought some \textcolor{red}{food} from the market at a store. & 
        \scriptsize When I grew up, there were \hl{many plants} in the \textcolor{red}{\hl{garden}}. & 
        \scriptsize The flower \textcolor{red}{\hl{garden}} was very beautiful. & 
        \scriptsize The \hl{garden} had \hl{many plants}. & 
        \scriptsize We had a great time \textcolor{blue}{in the end}! \\
        
        \small \textbf{MCSM+BART}: &
        \scriptsize Today we went to the park. & 
        \scriptsize There was a sign for \hl{location location location location}. & 
        \scriptsize It seemed that there were \textcolor{red}{two people} there. & 
        \scriptsize We left things there so we put them in the park. & 
        \scriptsize We decided to make a new sign with the \textcolor{red}{trees}. \\
        
        \small \textbf{Transformer+T5 (Ours)}: &
        \scriptsize I went to the flower shop to buy some flowers. & 
        \scriptsize \hl{I took a picture of} me and my friend \hl{while I was there}. & 
        \scriptsize \hl{There were a lot of} flowers for sale. & 
        \scriptsize \hl{There were a lot of} different colors. & 
        \scriptsize \hl{I took a picture of} my foot \hl{while I was there}. \\
        
        \small \textbf{Transformer+BART (Ours)}: &
        \scriptsize I went to the flower shop to buy some flowers and there was a huge variety of those. & 
        \scriptsize All the plants were very pretty. & 
        \scriptsize I bought some of the most beautiful margaritas. & 
        \scriptsize \textcolor{blue}{I also} bought some lilies and they were very expensive. & 
        \scriptsize I went back home and looked at my bought flowers. \\        
    \end{tabular}
    \vspace{-0.3cm}
    \caption{\centering Examples for qualitative comparison of our model against baselines. Words \hl{highlighted in yellow} are repetitive expressions, words \textcolor{red}{in red} represent content that is not relevant to the image sequence and words \textcolor{blue}{in blue} showcase some linking words that could potentially boost coherence.}
    \label{fig:comparison_with_others}
\end{figure*}

On Fig. \ref{fig:comparison_with_others} illustrates some visual examples from stories generated by two of our frameworks and the five models that took place in human and LLM evaluation. 
These examples are provided for the purpose of qualitative analysis and thus, we emphasize some distinctions. We color with red some visual inconsistencies, while we  highlight with yellow some nonsensical, repetitive and non-humane generations by the models. 
At the same time, we also include some blue colored words (or terms), which are mainly linking words, that in many cases increase cohesion and logical flow. 

\vspace{-0.1cm}
By examining these qualitative examples, we observe notable differences across the evaluated models in terms of coherence, fluency, and alignment with the visual content. 
A key observation is that our frameworks, especially Transf.+BART, exhibits improved logical flow, often utilizing linking words (highlighted in blue) that enhance readability and narrative continuity. 
In contrast, some external models generate outputs with disjointed sentence structures and repetitive phrasing (marked in yellow), leading to reduced narrative quality, something that, however, occurs in our Transf.+T5 model as well. 
Lastly, we see that our methods do not lapse into visual inconsistencies (highlighted with red), despite the fact that our approach is essentially text-only during story generation\footnote{The term text-only, corresponds to the fact that during story generation, our methods do not have any access to the actual visual embeddings of the input images. Our storytellers are only \say{seeing} textual information.}. 

\section{Conclusion}
\label{sec:conclusion}
\vspace{-0.4em}
In this work we showed that our framework demonstrates effectiveness in story generation maintaining, at the same time, simplicity and reproducibility. 
Despite the fact that we divide visual storytelling task into two distinct parts, separating visual from textual concepts in two stages, our approach manages to achieve strong capacity in keeping contextual consistency. 
By leveraging the idea that descriptive captions may capture the central information of a photo-stream, our methods accomplished semantically meaningful, coherent, but also visually grounded narratives, surpassing other state-of-the-art baselines in numerous types of evaluations. 
We therefore, claim that as captioning systems progress and narration models become more robust, utilizing simple and cost-effective approaches such as the one presented, could result in high-quality visual storytelling, without the need of complicated end-to-end architectures.  

However, despite these improvements of our method, certain challenges remain, such as occasional redundancy or simplistic sentence structure, suggesting room for further refinements in our methodology.
Especially when compared against next-generation multimodal LLMs, such as StoryLLaVA, we can certify that the latter produces, lengthier and more realistic stories that enclose, world-knowledge, an aspect that our system is currently lacking.  
To that end, for future work we aim to explore the integration of Retrieval-Augmented Generation (RAG) or external knowledge graphs to enrich our model’s access to real-time insights. 
By incorporating such mechanisms, we can enhance the factual grounding and contextual depth of generated stories, addressing current limitations in implicit knowledge representation.
Ultimately, this work underscores the importance of bridging vision and language more effectively, paving the way for future advancements in visual storytelling.

\section*{Acknowledgments}
I would like to thank Utrecht University, my supervisor Prof. Albert Gatt and the PhD student Yingjin Song for giving me the chance to complete this MSc Thesis project. Their guidance and support throughout the research was instrumental. 
I also acknowledge the time that my participants dedicated for making the evaluation procedure more comprehensive. Lastly, I would like to thank Frank Ferraro (University of Maryland, Baltimore Country) for sharing the VIST data with our team.


\bibliographystyle{acl_natbib}
\bibliography{refs}

\begin{thebibliography}{91}
\expandafter\ifx\csname natexlab\endcsname\relax\def\natexlab#1{#1}\fi

\bibitem[{Anderson et~al.(2016)Anderson, Fernando, Johnson, and Gould}]{anderson2016spicesemanticpropositionalimage}
Peter Anderson, Basura Fernando, Mark Johnson, and Stephen Gould. 2016.
\newblock \href {http://arxiv.org/abs/1607.08822} {Spice: Semantic propositional image caption evaluation}.

\bibitem[{Anderson et~al.(2018)Anderson, He, Buehler, Teney, Johnson, Gould, and Zhang}]{anderson2018bottomuptopdownattentionimage}
Peter Anderson, Xiaodong He, Chris Buehler, Damien Teney, Mark Johnson, Stephen Gould, and Lei Zhang. 2018.
\newblock \href {http://arxiv.org/abs/1707.07998} {Bottom-up and top-down attention for image captioning and visual question answering}.

\bibitem[{Bai et~al.(2023)Bai, Yang, Bai, Wang, Zhang, Lin, Wang, Zhou, and Zhou}]{bai2023touchstoneevaluatingvisionlanguagemodels}
Shuai Bai, Shusheng Yang, Jinze Bai, Peng Wang, Xingxuan Zhang, Junyang Lin, Xinggang Wang, Chang Zhou, and Jingren Zhou. 2023.
\newblock \href {http://arxiv.org/abs/2308.16890} {Touchstone: Evaluating vision-language models by language models}.

\bibitem[{Banerjee and Lavie(2005)}]{banerjee-lavie-2005-meteor}
Satanjeev Banerjee and Alon Lavie. 2005.
\newblock \href {https://aclanthology.org/W05-0909/} {{METEOR}: An automatic metric for {MT} evaluation with improved correlation with human judgments}.
\newblock In \emph{Proceedings of the {ACL} Workshop on Intrinsic \& Extrinsic Evaluation Measures for Machine Translation and Summarization}, pages 65--72, Ann Arbor, Michigan. Association for Computational Linguistics.

\bibitem[{Chen et~al.(2022{\natexlab{a}})Chen, Huang, Takamura, and Nakayama}]{chen2021commonsenseknowledgeawareconcept}
Hong Chen, Yifei Huang, Hiroya Takamura, and Hideki Nakayama. 2022{\natexlab{a}}.
\newblock \href {http://arxiv.org/abs/2102.02963} {Commonsense knowledge aware concept selection for diverse and informative visual storytelling}.

\bibitem[{Chen et~al.(2022{\natexlab{b}})Chen, Liu, and Niu}]{9797749}
Wei Chen, Xuefeng Liu, and Jianwei Niu. 2022{\natexlab{b}}.
\newblock \href {https://doi.org/10.1109/TCSVT.2022.3183648} {Sentistory: A multi-layered sentiment-aware generative model for visual storytelling}.
\newblock \emph{IEEE Transactions on Circuits and Systems for Video Technology}.

\bibitem[{Chen et~al.(2024)Chen, Li, Su, Zhu, Li, Ji, and Liu}]{chen2024tarnvisttopicawarereinforcement}
Weiran Chen, Xin Li, Jiaqi Su, Guiqian Zhu, Ying Li, Yi~Ji, and Chunping Liu. 2024.
\newblock \href {http://arxiv.org/abs/2403.11550} {Tarn-vist: Topic aware reinforcement network for visual storytelling}.

\bibitem[{Chhun et~al.(2024)Chhun, Suchanek, and Clavel}]{chhun2024language}
Cyril Chhun, Fabian~M Suchanek, and Chlo{\'e} Clavel. 2024.
\newblock Do language models enjoy their own stories? prompting large language models for automatic story evaluation.
\newblock \emph{Transactions of the Association for Computational Linguistics}, 12:1122--1142.

\bibitem[{Devlin et~al.(2019)Devlin, Chang, Lee, and Toutanova}]{devlin-etal-2019-bert}
Jacob Devlin, Ming-Wei Chang, Kenton Lee, and Kristina Toutanova. 2019.
\newblock \href {https://doi.org/10.18653/v1/N19-1423} {{BERT}: Pre-training of deep bidirectional transformers for language understanding}.
\newblock In \emph{Proceedings of the 2019 Conference of the North {A}merican Chapter of the Association for Computational Linguistics: Human Language Technologies}, Minneapolis, Minnesota. Association for Computational Linguistics.

\bibitem[{Devlin et~al.(2015)Devlin, Gupta, Girshick, Mitchell, and Zitnick}]{devlin2015exploringnearestneighborapproaches}
Jacob Devlin, Saurabh Gupta, Ross Girshick, Margaret Mitchell, and C.~Lawrence Zitnick. 2015.
\newblock \href {http://arxiv.org/abs/1505.04467} {Exploring nearest neighbor approaches for image captioning}.

\bibitem[{Dosovitskiy et~al.(2021)Dosovitskiy, Beyer, Kolesnikov, Weissenborn, Zhai, Unterthiner, Dehghani, Minderer, Heigold, Gelly, Uszkoreit, and Houlsby}]{dosovitskiy2021imageworth16x16words}
Alexey Dosovitskiy, Lucas Beyer, Alexander Kolesnikov, Dirk Weissenborn, Xiaohua Zhai, Thomas Unterthiner, Mostafa Dehghani, Matthias Minderer, Georg Heigold, Sylvain Gelly, Jakob Uszkoreit, and Neil Houlsby. 2021.
\newblock \href {http://arxiv.org/abs/2010.11929} {An image is worth 16x16 words: Transformers for image recognition at scale}.

\bibitem[{Fan et~al.(2022)Fan, Wang, Gu, and Liu}]{10.1145/3469877.3490604}
Ruichao Fan, Hanli Wang, Jinjing Gu, and Xianhui Liu. 2022.
\newblock \href {https://doi.org/10.1145/3469877.3490604} {Visual storytelling with hierarchical bert semantic guidance}.
\newblock In \emph{Proceedings of the 3rd ACM International Conference on Multimedia in Asia}, New York, USA. Association for Computing Machinery.

\bibitem[{Fang et~al.(2021)Fang, Zeng, Liu, Bo, Dong, and Chen}]{fang2021transformerbasedconditionalvariationalautoencoder}
Le~Fang, Tao Zeng, Chaochun Liu, Liefeng Bo, Wen Dong, and Changyou Chen. 2021.
\newblock \href {http://arxiv.org/abs/2101.00828} {Transformer-based conditional variational autoencoder for controllable story generation}.

\bibitem[{Farhadi et~al.(2010)Farhadi, Hejrati, Sadeghi, Young, Rashtchian, Hockenmaier, and Forsyth}]{farhadi2010every}
Ali Farhadi, Mohsen Hejrati, Mohammad~Amin Sadeghi, Peter Young, Cyrus Rashtchian, Julia Hockenmaier, and David Forsyth. 2010.
\newblock Every picture tells a story: Generating sentences from images.
\newblock In \emph{Computer Vision--ECCV 2010: 11th European Conference on Computer Vision, Heraklion, Crete, Greece, Proceedings, Part IV 11}, pages 15--29. Springer.

\bibitem[{Gerv{\'a}s(2009)}]{gervas2009computational}
Pablo Gerv{\'a}s. 2009.
\newblock Computational approaches to storytelling and creativity.
\newblock \emph{AI Magazine}, 30(3):49--49.

\bibitem[{Gonzalez-Rico and Fuentes-Pineda(2018)}]{gonzalezrico2018contextualizetellneuralvisual}
Diana Gonzalez-Rico and Gibran Fuentes-Pineda. 2018.
\newblock \href {http://arxiv.org/abs/1806.00738} {Contextualize, show and tell: A neural visual storyteller}.

\bibitem[{Guan et~al.(2020)Guan, Huang, Zhao, Zhu, and Huang}]{guan2020knowledgeenhancedpretrainingmodelcommonsense}
Jian Guan, Fei Huang, Zhihao Zhao, Xiaoyan Zhu, and Minlie Huang. 2020.
\newblock \href {http://arxiv.org/abs/2001.05139} {A knowledge-enhanced pretraining model for commonsense story generation}.

\bibitem[{Han et~al.(2020)Han, Long, Luo, Wang, and Poon}]{han2020victrvisualinformationcaptured}
Soyeon~Caren Han, Siqu Long, Siwen Luo, Kunze Wang, and Josiah Poon. 2020.
\newblock \href {http://arxiv.org/abs/2010.03182} {Victr: Visual information captured text representation for text-to-image multimodal tasks}.

\bibitem[{He et~al.(2015)He, Zhang, Ren, and Sun}]{he2015deepresiduallearningimage}
Kaiming He, Xiangyu Zhang, Shaoqing Ren, and Jian Sun. 2015.
\newblock \href {http://arxiv.org/abs/1512.03385} {Deep residual learning for image recognition}.

\bibitem[{He et~al.(2020)He, Liao, Tavakoli, Yang, Rosenhahn, and Pugeault}]{he2020imagecaptioningimagetransformer}
Sen He, Wentong Liao, Hamed~R. Tavakoli, Michael Yang, Bodo Rosenhahn, and Nicolas Pugeault. 2020.
\newblock \href {http://arxiv.org/abs/2004.14231} {Image captioning through image transformer}.

\bibitem[{Herdade et~al.(2020)Herdade, Kappeler, Boakye, and Soares}]{herdade2020imagecaptioningtransformingobjects}
Simao Herdade, Armin Kappeler, Kofi Boakye, and Joao Soares. 2020.
\newblock \href {http://arxiv.org/abs/1906.05963} {Image captioning: Transforming objects into words}.

\bibitem[{Hong et~al.(2023)Hong, Sayeed, Mehra, Demberg, and Schiele}]{hong2023visualwritingpromptscharactergrounded}
Xudong Hong, Asad Sayeed, Khushboo Mehra, Vera Demberg, and Bernt Schiele. 2023.
\newblock \href {http://arxiv.org/abs/2301.08571} {Visual writing prompts: Character-grounded story generation with curated image sequences}.

\bibitem[{Hong et~al.(2020)Hong, Shetty, Sayeed, Mehra, Demberg, and Schiele}]{hong-etal-2020-diverse}
Xudong Hong, Rakshith Shetty, Asad Sayeed, Khushboo Mehra, Vera Demberg, and Bernt Schiele. 2020.
\newblock \href {https://doi.org/10.18653/v1/2020.conll-1.34} {Diverse and relevant visual storytelling with scene graph embeddings}.
\newblock In \emph{Proceedings of the 24th Conference on Computational Natural Language Learning}. Association for Computational Linguistics.

\bibitem[{Hsu et~al.(2019{\natexlab{a}})Hsu, Chen, Hsu, Li, Lin, Huang, and Ku}]{hsu2019knowledgeenrichedvisualstorytelling}
Chao-Chun Hsu, Zi-Yuan Chen, Chi-Yang Hsu, Chih-Chia Li, Tzu-Yuan Lin, Ting-Hao~'Kenneth' Huang, and Lun-Wei Ku. 2019{\natexlab{a}}.
\newblock \href {http://arxiv.org/abs/1912.01496} {Knowledge-enriched visual storytelling}.

\bibitem[{Hsu et~al.(2021)Hsu, Chu, Huang, and Ku}]{hsu2021plotreworkmodelingstorylines}
Chi-Yang Hsu, Yun-Wei Chu, Ting-Hao~'Kenneth' Huang, and Lun-Wei Ku. 2021.
\newblock \href {http://arxiv.org/abs/2105.06950} {Plot and rework: Modeling storylines for visual storytelling}.

\bibitem[{Hsu et~al.(2019{\natexlab{b}})Hsu, Huang, Hsu, and Huang}]{hsu2019visualstorypostediting}
Ting-Yao Hsu, Chieh-Yang Huang, Yen-Chia Hsu, and Ting-Hao~'Kenneth' Huang. 2019{\natexlab{b}}.
\newblock \href {http://arxiv.org/abs/1906.01764} {Visual story post-editing}.

\bibitem[{Hu et~al.(2020)Hu, Cheng, Gan, Liu, Gao, and Neubig}]{hu2020makesgoodstorydesigning}
Junjie Hu, Yu~Cheng, Zhe Gan, Jingjing Liu, Jianfeng Gao, and Graham Neubig. 2020.
\newblock \href {http://arxiv.org/abs/1909.05316} {What makes a good story? designing composite rewards for visual storytelling}.

\bibitem[{Huang et~al.(2019)Huang, Wang, Chen, and Wei}]{huang2019attentionattentionimagecaptioning}
Lun Huang, Wenmin Wang, Jie Chen, and Xiao-Yong Wei. 2019.
\newblock \href {http://arxiv.org/abs/1908.06954} {Attention on attention for image captioning}.

\bibitem[{Huang et~al.(2016)Huang, Ferraro, Mostafazadeh, Misra, Agrawal, Devlin, Girshick, He, Kohli, Batra, Zitnick, Parikh, Vanderwende, Michel, and Margaret}]{huang-etal-2016-visual}
Ting-Hao~Kenneth Huang, Francis Ferraro, Nasrin Mostafazadeh, Ishan Misra, Aishwarya Agrawal, Jacob Devlin, Ross Girshick, Xiaodong He, Pushmeet Kohli, Dhruv Batra, C.~Lawrence Zitnick, Devi Parikh, Lucy Vanderwende, Galley Michel, and Mitchell Margaret. 2016.
\newblock \href {https://doi.org/10.18653/v1/N16-1147} {Visual storytelling}.
\newblock In \emph{Proceedings of the 2016 Conference of North {A}merican Chapter of the Association for Computational Linguistics: Human Language Technologies}, San Diego, California. Association for Computational Linguistics.

\bibitem[{Jain et~al.(2017)Jain, Agrawal, Mishra, Sukhwani, Laha, and Sankaranarayanan}]{jain2017storygenerationsequenceindependent}
Parag Jain, Priyanka Agrawal, Abhijit Mishra, Mohak Sukhwani, Anirban Laha, and Karthik Sankaranarayanan. 2017.
\newblock \href {http://arxiv.org/abs/1707.05501} {Story generation from sequence of independent short descriptions}.

\bibitem[{Jung et~al.(2020)Jung, Kim, Woo, Kim, Kim, and Kweon}]{jung2020hide}
Yunjae Jung, Dahun Kim, Sanghyun Woo, Kyungsu Kim, Sungjin Kim, and In~So Kweon. 2020.
\newblock Hide-and-tell: Learning to bridge photo streams for visual storytelling.
\newblock In \emph{Proceedings of the AAAI Conference on Artificial Intelligence}, 07, pages 11213--11220.

\bibitem[{Kim(2015)}]{kim2015understanding}
Jeong-Hee Kim. 2015.
\newblock \emph{Understanding narrative inquiry: The crafting and analysis of stories as research}.
\newblock Sage publications.

\bibitem[{Kim et~al.(2019)Kim, Heo, Son, Park, and Zhang}]{kim2019glacnetglocalattention}
Taehyeong Kim, Min-Oh Heo, Seonil Son, Kyoung-Wha Park, and Byoung-Tak Zhang. 2019.
\newblock \href {http://arxiv.org/abs/1805.10973} {Glac net: Glocal attention cascading networks for multi-image cued story generation}.

\bibitem[{Kingma and Ba(2017)}]{kingma2017adammethodstochasticoptimization}
Diederik~P. Kingma and Jimmy Ba. 2017.
\newblock \href {http://arxiv.org/abs/1412.6980} {Adam: A method for stochastic optimization}.

\bibitem[{Krishna et~al.(2016)Krishna, Zhu, Groth, Johnson, Hata, Kravitz, Chen, Kalantidis, Li, Shamma, Bernstein, and Li}]{krishna2016visualgenomeconnectinglanguage}
Ranjay Krishna, Yuke Zhu, Oliver Groth, Justin Johnson, Kenji Hata, Joshua Kravitz, Stephanie Chen, Yannis Kalantidis, Li-Jia Li, David~A. Shamma, Michael~S. Bernstein, and Fei-Fei Li. 2016.
\newblock \href {http://arxiv.org/abs/1602.07332} {Visual genome: Connecting language and vision using crowdsourced dense image annotations}.

\bibitem[{Lewis et~al.(2020)Lewis, Liu, Goyal, Ghazvininejad, Mohamed, Levy, Stoyanov, and Zettlemoyer}]{lewis-etal-2020-bart}
Mike Lewis, Yinhan Liu, Naman Goyal, Marjan Ghazvininejad, Abdelrahman Mohamed, Omer Levy, Veselin Stoyanov, and Luke Zettlemoyer. 2020.
\newblock \href {https://doi.org/10.18653/v1/2020.acl-main.703} {{BART}: Denoising sequence-to-sequence pre-training for natural language generation, translation, and comprehension}.
\newblock In \emph{Proceedings of the 58th Annual Meeting of the Association for Computational Linguistics}, Online. Association for Computational Linguistics.

\bibitem[{Liang et~al.(2024)Liang, Zhang, Cao, Wang, Ding, Yang, Vodrahalli, He, Smith, Yin, McFarland, and Zou}]{liang2023largelanguagemodelsprovide}
Weixin Liang, Yuhui Zhang, Hancheng Cao, Binglu Wang, Daisy Ding, Xinyu Yang, Kailas Vodrahalli, Siyu He, Daniel Smith, Yian Yin, Daniel McFarland, and James Zou. 2024.
\newblock \href {http://arxiv.org/abs/2310.01783} {Can large language models provide useful feedback on research papers? a large-scale empirical analysis}.

\bibitem[{Lin(2004)}]{lin2004rouge}
Chin-Yew Lin. 2004.
\newblock Rouge: A package for automatic evaluation of summaries.
\newblock In \emph{Text summarization branches out}, pages 74--81.

\bibitem[{Lin et~al.(2014)Lin, Maire, Belongie, Bourdev, Girshick, Hays, Perona, Ramanan, Doll{\'{a}}r, and Zitnick}]{DBLP:journals/corr/LinMBHPRDZ14}
Tsung{-}Yi Lin, Michael Maire, Serge~J. Belongie, Lubomir~D. Bourdev, Ross~B. Girshick, James Hays, Pietro Perona, Deva Ramanan, Piotr Doll{\'{a}}r, and C.~Lawrence Zitnick. 2014.
\newblock \href {http://arxiv.org/abs/1405.0312} {Microsoft {COCO:} common objects in context}.
\newblock \emph{CoRR}, abs/1405.0312.

\bibitem[{Liu and Keller(2023)}]{liu2023detectinggroundingimportantcharacters}
Danyang Liu and Frank Keller. 2023.
\newblock \href {http://arxiv.org/abs/2303.17647} {Detecting and grounding important characters in visual stories}.

\bibitem[{Liu et~al.(2024)Liu, Lapata, and Keller}]{liu2024generatingvisualstoriesgrounded}
Danyang Liu, Mirella Lapata, and Frank Keller. 2024.
\newblock \href {http://arxiv.org/abs/2409.13555} {Generating visual stories with grounded and coreferent characters}.

\bibitem[{Liu et~al.(2023{\natexlab{a}})Liu, Yang, Chang, Wang, Zheng, Jiang, Wang, Xie, and Wu}]{liu2023aog}
Hanqing Liu, Jiacheng Yang, Chia-Hao Chang, Wei Wang, Hai-Tao Zheng, Yong Jiang, Hui Wang, Rui Xie, and Wei Wu. 2023{\natexlab{a}}.
\newblock Aog-lstm: An adaptive attention neural network for visual storytelling.
\newblock \emph{Neurocomputing}, 552:126486.

\bibitem[{Liu et~al.(2023{\natexlab{b}})Liu, Li, Wu, and Lee}]{liu2023visualinstructiontuning}
Haotian Liu, Chunyuan Li, Qingyang Wu, and Yong~Jae Lee. 2023{\natexlab{b}}.
\newblock \href {http://arxiv.org/abs/2304.08485} {Visual instruction tuning}.

\bibitem[{Liu and Singh(2004)}]{liu2004conceptnet}
Hugo Liu and Push Singh. 2004.
\newblock Conceptnet—a practical commonsense reasoning tool-kit.
\newblock \emph{BT technology journal}, 22(4):211--226.

\bibitem[{Liu et~al.(2017)Liu, Fu, Mei, and Chen}]{10.55556/3298239.3298450}
Yu~Liu, Jianlong Fu, Tao Mei, and Chang~Wen Chen. 2017.
\newblock Let your photos talk: generating narrative paragraph for photo stream via bidirectional attention recurrent neural networks.
\newblock In \emph{Proceedings of the Thirty-First AAAI Conference on Artificial Intelligence}, AAAI'17, page 1445–1452. AAAI Press.

\bibitem[{Loshchilov and Hutter(2019)}]{loshchilov2019decoupledweightdecayregularization}
Ilya Loshchilov and Frank Hutter. 2019.
\newblock \href {http://arxiv.org/abs/1711.05101} {Decoupled weight decay regularization}.

\bibitem[{Lu et~al.(2019)Lu, Batra, Parikh, and Lee}]{lu2019vilbertpretrainingtaskagnosticvisiolinguistic}
Jiasen Lu, Dhruv Batra, Devi Parikh, and Stefan Lee. 2019.
\newblock \href {http://arxiv.org/abs/1908.02265} {Vilbert: Pretraining task-agnostic visiolinguistic representations for vision-and-language tasks}.

\bibitem[{Lu et~al.(2017)Lu, Xiong, Parikh, and Socher}]{lu2017knowinglookadaptiveattention}
Jiasen Lu, Caiming Xiong, Devi Parikh, and Richard Socher. 2017.
\newblock \href {http://arxiv.org/abs/1612.01887} {Knowing when to look: Adaptive attention via a visual sentinel for image captioning}.

\bibitem[{Mokady et~al.(2021)Mokady, Hertz, and Bermano}]{DBLP:journals/corr/abs-2111-09734}
Ron Mokady, Amir Hertz, and Amit~H. Bermano. 2021.
\newblock \href {http://arxiv.org/abs/2111.09734} {Clipcap: {CLIP} prefix for image captioning}.
\newblock \emph{CoRR}, abs/2111.09734.

\bibitem[{Ning et~al.(2023)Ning, Xie, Chen, Song, Yuan, Tian, Ye, and Yuan}]{ning2023albumstorytellingiterativestoryaware}
Munan Ning, Yujia Xie, Dongdong Chen, Zeyin Song, Lu~Yuan, Yonghong Tian, Qixiang Ye, and Li~Yuan. 2023.
\newblock \href {http://arxiv.org/abs/2305.12943} {Album storytelling with iterative story-aware captioning and large language models}.

\bibitem[{OpenAI(2023)}]{openai2024gpt4technicalreport}
OpenAI. 2023.
\newblock \href {http://arxiv.org/abs/2303.08774} {Gpt-4 technical report}.

\bibitem[{Papineni et~al.(2002)Papineni, Roukos, Ward, and Zhu}]{papineni2002bleu}
Kishore Papineni, Salim Roukos, Todd Ward, and Wei-Jing Zhu. 2002.
\newblock Bleu: a method for automatic evaluation of machine translation.
\newblock In \emph{Proceedings of the 40th annual meeting of the Association for Computational Linguistics}, pages 311--318.

\bibitem[{Park and Kim(2015)}]{NIPS2015_17e62166}
Cesc~C Park and Gunhee Kim. 2015.
\newblock \href {https://proceedings.neurips.cc/paper_files/paper/2015/file/17e62166fc8586dfa4d1bc0e1742c08b-Paper.pdf} {Expressing an image stream with a sequence of natural sentences}.
\newblock In \emph{Advances in Neural Information Processing Systems}, volume~28. Curran Associates, Inc.

\bibitem[{Prabhumoye et~al.(2019)Prabhumoye, Chandu, Salakhutdinov, and Black}]{prabhumoye2019mywaytellingstory}
Shrimai Prabhumoye, Khyathi~Raghavi Chandu, Ruslan Salakhutdinov, and Alan~W Black. 2019.
\newblock \href {http://arxiv.org/abs/1906.06401} {"my way of telling a story": Persona based grounded story generation}.

\bibitem[{Qi et~al.(2021)Qi, Qin, Huang, Shen, Yang, and Luo}]{10.1145/3474085.3475236}
Mengshi Qi, Jie Qin, Di~Huang, Zhiqiang Shen, Yi~Yang, and Jiebo Luo. 2021.
\newblock \href {https://doi.org/10.1145/3474085.3475236} {Latent memory-augmented graph transformer for visual storytelling}.
\newblock In \emph{Proceedings of the 29th ACM International Conference on Multimedia}, page 4892–4901, New York, NY, USA. Association for Computing Machinery.

\bibitem[{Radford(2018)}]{radford2018improving}
Alec Radford. 2018.
\newblock Improving language understanding by generative pre-training.
\newblock \emph{Open AI}.

\bibitem[{Radford et~al.(2021)Radford, Kim, Hallacy, Ramesh, Goh, Agarwal, Sastry, Askell, Mishkin, Clark et~al.}]{radford2021learning}
Alec Radford, Jong~Wook Kim, Chris Hallacy, Aditya Ramesh, Gabriel Goh, Sandhini Agarwal, Girish Sastry, Amanda Askell, Pamela Mishkin, Jack Clark, et~al. 2021.
\newblock Learning transferable visual models from natural language supervision.
\newblock In \emph{International conference on machine learning}, pages 8748--8763. PMLR.

\bibitem[{Radford et~al.(2019)Radford, Wu, Child, Luan, Amodei, Sutskever et~al.}]{radford2019language}
Alec Radford, Jeffrey Wu, Rewon Child, David Luan, Dario Amodei, Ilya Sutskever, et~al. 2019.
\newblock Language models are unsupervised multitask learners.
\newblock \emph{OpenAI blog}, 1(8):9.

\bibitem[{Raffel et~al.(2020)Raffel, Shazeer, Roberts, Lee, Narang, Matena, Zhou, Li, and Liu}]{JMLR:v21:20-074}
Colin Raffel, Noam Shazeer, Adam Roberts, Katherine Lee, Sharan Narang, Michael Matena, Yanqi Zhou, Wei Li, and Peter~J. Liu. 2020.
\newblock \href {http://jmlr.org/papers/v21/20-074.html} {Exploring the limits of transfer learning with a unified text-to-text transformer}.
\newblock \emph{Journal of Machine Learning Research}, 21(140):1--67.

\bibitem[{Ravi et~al.(2021)Ravi, Kafle, Cohen, Brandt, and Kapadia}]{Ravi_2021_ICCV}
Hareesh Ravi, Kushal Kafle, Scott Cohen, Jonathan Brandt, and Mubbasir Kapadia. 2021.
\newblock Aesop: Abstract encoding of stories, objects, and pictures.
\newblock In \emph{Proceedings of the IEEE/CVF International Conference on Computer Vision (ICCV)}, pages 2052--2063.

\bibitem[{Rennie et~al.(2017)Rennie, Marcheret, Mroueh, Ross, and Goel}]{rennie2017selfcriticalsequencetrainingimage}
Steven~J. Rennie, Etienne Marcheret, Youssef Mroueh, Jarret Ross, and Vaibhava Goel. 2017.
\newblock \href {http://arxiv.org/abs/1612.00563} {Self-critical sequence training for image captioning}.

\bibitem[{Sellam et~al.(2020)Sellam, Das, and Parikh}]{sellam-etal-2020-bleurt}
Thibault Sellam, Dipanjan Das, and Ankur Parikh. 2020.
\newblock \href {https://doi.org/10.18653/v1/2020.acl-main.704} {{BLEURT}: Learning robust metrics for text generation}.
\newblock In \emph{Proceedings of the 58th Annual Meeting of the Association for Computational Linguistics}, pages 7881--7892, Online. Association for Computational Linguistics.

\bibitem[{Shannon(1948)}]{shannon1948mathematical}
Claude~Elwood Shannon. 1948.
\newblock A mathematical theory of communication.
\newblock \emph{The Bell system technical journal}, 27(3):379--423.

\bibitem[{Smilevski et~al.(2018)Smilevski, Lalkovski, and Madjarov}]{Smilevski_2018}
Marko Smilevski, Ilija Lalkovski, and Gjorgji Madjarov. 2018.
\newblock \href {https://doi.org/10.1007/978-3-030-00825-3_13} {\emph{Stories for Images-in-Sequence by Using Visual and Narrative Components}}, page 148–159. Springer International Publishing.

\bibitem[{Song et~al.(2024)Song, Paperno, and Gatt}]{song2024contextawarevisualstorytellingvisual}
Yingjin Song, Denis Paperno, and Albert Gatt. 2024.
\newblock \href {http://arxiv.org/abs/2408.06259} {Context-aware visual storytelling with visual prefix tuning and contrastive learning}.

\bibitem[{Su et~al.(2020)Su, Dai, Guerin, and Zhou}]{su2020berthlstmsberthierarchicallstms}
Jing Su, Qingyun Dai, Frank Guerin, and Mian Zhou. 2020.
\newblock \href {http://arxiv.org/abs/2012.02128} {Bert-hlstms: Bert and hierarchical lstms for visual storytelling}.

\bibitem[{Su et~al.(2022)Su, Lan, Wang, Yogatama, Kong, and Collier}]{su2022contrastiveframeworkneuraltext}
Yixuan Su, Tian Lan, Yan Wang, Dani Yogatama, Lingpeng Kong, and Nigel Collier. 2022.
\newblock \href {http://arxiv.org/abs/2202.06417} {A contrastive framework for neural text generation}.

\bibitem[{Surikuchi et~al.(2023)Surikuchi, Pezzelle, and Fernández}]{surikuchi2023groovistmetricgroundingobjects}
Aditya~K Surikuchi, Sandro Pezzelle, and Raquel Fernández. 2023.
\newblock \href {http://arxiv.org/abs/2310.17770} {Groovist: A metric for grounding objects in visual storytelling}.

\bibitem[{Tanaka and Aihara(2015)}]{tanaka-ishii-aihara-2015-computational}
Kumiko Tanaka and Shunsuke Aihara. 2015.
\newblock \href {https://doi.org/10.1162/COLI_a_00228} {Computational constancy measures of {T}exts{---}{Y}ule`s k and r{\'e}nyi`s entropy}.
\newblock \emph{Computational Linguistics}, 41(3).

\bibitem[{Vaswani et~al.(2017)Vaswani, Shazeer, Parmar, Uszkoreit, Jones, Gomez, Kaiser, and Polosukhin}]{DBLP:journals/corr/VaswaniSPUJGKP17}
Ashish Vaswani, Noam Shazeer, Niki Parmar, Jakob Uszkoreit, Llion Jones, Aidan~N. Gomez, Lukasz Kaiser, and Illia Polosukhin. 2017.
\newblock \href {http://arxiv.org/abs/1706.03762} {Attention is all you need}.
\newblock \emph{CoRR}, abs/1706.03762.

\bibitem[{Vedantam et~al.(2015)Vedantam, Zitnick, and Parikh}]{vedantam2015ciderconsensusbasedimagedescription}
Ramakrishna Vedantam, C.~Lawrence Zitnick, and Devi Parikh. 2015.
\newblock \href {http://arxiv.org/abs/1411.5726} {Cider: Consensus-based image description evaluation}.

\bibitem[{Vinyals et~al.(2015)Vinyals, Toshev, Bengio, and Erhan}]{vinyals2015tellneuralimagecaption}
Oriol Vinyals, Alexander Toshev, Samy Bengio, and Dumitru Erhan. 2015.
\newblock \href {http://arxiv.org/abs/1411.4555} {Show and tell: A neural image caption generator}.

\bibitem[{Wang et~al.(2022)Wang, Han, and Poon}]{wang2022rovistlearningrobustmetricsvisual}
Eileen Wang, Caren Han, and Josiah Poon. 2022.
\newblock \href {http://arxiv.org/abs/2205.03774} {Rovist:learning robust metrics for visual storytelling}.

\bibitem[{Wang et~al.(2024)Wang, Han, and Poon}]{wang2024scovistsocialinteractioncommonsense}
Eileen Wang, Soyeon~Caren Han, and Josiah Poon. 2024.
\newblock \href {http://arxiv.org/abs/2402.00319} {Sco-vist: Social interaction commonsense knowledge-based visual storytelling}.

\bibitem[{Wang et~al.(2018{\natexlab{a}})Wang, Fu, Tang, Li, and Mei}]{10.5555/3504035.3504941}
Jing Wang, Jianlong Fu, Jinhui Tang, Zechao Li, and Tao Mei. 2018{\natexlab{a}}.
\newblock Show, reward and tell: automatic generation of narrative paragraph from photo stream by adversarial training.
\newblock In \emph{Proceedings of the Thirty-Second AAAI Conference on Artificial Intelligence and Thirtieth Innovative Applications of Artificial Intelligence Conference and Eighth AAAI Symposium on Educational Advances in AI}. AAAI Press.

\bibitem[{Wang et~al.(2020)Wang, Wei, Li, Zhang, and Huang}]{Wang_Wei_Li_Zhang_Huang_2020}
Ruize Wang, Zhongyu Wei, Piji Li, Qi~Zhang, and Xuanjing Huang. 2020.
\newblock \href {https://doi.org/10.1609/aaai.v34i05.6455} {Storytelling from an image stream using scene graphs}.
\newblock \emph{Proceedings of the AAAI Conference on Artificial Intelligence}, 34(05):9185--9192.

\bibitem[{Wang et~al.(2018{\natexlab{b}})Wang, Chen, Wang, and Wang}]{wang2018metricsperfectadversarialreward}
Xin Wang, Wenhu Chen, Yuan-Fang Wang, and William~Yang Wang. 2018{\natexlab{b}}.
\newblock \href {http://arxiv.org/abs/1804.09160} {No metrics are perfect: Adversarial reward learning for visual storytelling}.

\bibitem[{Wang et~al.(2023)Wang, Zhou, Lu, and Li}]{10.1145/3581783.3612179}
Yuechen Wang, Wengang Zhou, Zhenbo Lu, and Houqiang Li. 2023.
\newblock \href {https://doi.org/10.1145/3581783.3612179} {Text-only training for visual storytelling}.
\newblock In \emph{Proceedings of the 31st ACM International Conference on Multimedia}, New York, NY, USA. Association for Computing Machinery.

\bibitem[{Wu et~al.(2023)Wu, Koo, Blum, Black, Kao, Scalzo, and Kurtz}]{wu2023comparativestudyopensourcelarge}
Sean Wu, Michael Koo, Lesley Blum, Andy Black, Liyo Kao, Fabien Scalzo, and Ira Kurtz. 2023.
\newblock \href {http://arxiv.org/abs/2308.04709} {A comparative study of open-source large language models, gpt-4 and claude 2: Multiple-choice test taking in nephrology}.

\bibitem[{Xu et~al.(2021)Xu, Yang, Li, Shen, Ao, and Xu}]{xu2021imagine}
Chunpu Xu, Min Yang, Chengming Li, Ying Shen, Xiang Ao, and Ruifeng Xu. 2021.
\newblock Imagine, reason and write: Visual storytelling with graph knowledge and relational reasoning.
\newblock In \emph{Proceedings of the AAAI Conference on Artificial Intelligence}, volume~35, pages 3022--3029.

\bibitem[{Xu et~al.(2016)Xu, Ba, Kiros, Cho, Courville, Salakhutdinov, Zemel, and Bengio}]{xu2016showattendtellneural}
Kelvin Xu, Jimmy Ba, Ryan Kiros, Kyunghyun Cho, Aaron Courville, Ruslan Salakhutdinov, Richard Zemel, and Yoshua Bengio. 2016.
\newblock \href {http://arxiv.org/abs/1502.03044} {Show, attend and tell: Neural image caption generation with visual attention}.

\bibitem[{Xu et~al.(2020)Xu, Patwary, Shoeybi, Puri, Fung, Anandkumar, and Catanzaro}]{xu-etal-2020-megatron}
Peng Xu, Mostofa Patwary, Mohammad Shoeybi, Raul Puri, Pascale Fung, Anima Anandkumar, and Bryan Catanzaro. 2020.
\newblock \href {https://doi.org/10.18653/v1/2020.emnlp-main.226} {{MEGATRON}-{CNTRL}: Controllable story generation with external knowledge using large-scale language models}.
\newblock In \emph{Proceedings of the 2020 Conference on Empirical Methods in Natural Language Processing (EMNLP)}, Online. Association for Computational Linguistics.

\bibitem[{Yang and Jin(2023)}]{yang-jin-2023-attractive}
Dingyi Yang and Qin Jin. 2023.
\newblock \href {https://doi.org/10.18653/v1/2023.acl-long.619} {Attractive storyteller: Stylized visual storytelling with unpaired text}.
\newblock In \emph{Proceedings of the 61st Annual Meeting of the Association for Computational Linguistics (Volume 1: Long Papers)}, pages 11053--11066, Toronto, Canada. Association for Computational Linguistics.

\bibitem[{Yang et~al.(2025)Yang, Xiao, Huang, and Zhong}]{yang-etal-2025-storyllava}
Li~Yang, Zhiding Xiao, Wenxin Huang, and Xian Zhong. 2025.
\newblock \href {https://aclanthology.org/2025.coling-main.266/} {{S}tory{LL}a{VA}: Enhancing visual storytelling with multi-modal large language models}.
\newblock In \emph{Proceedings of the 31st International Conference on Computational Linguistics}, pages 3936--3951, Abu Dhabi, UAE. Association for Computational Linguistics.

\bibitem[{Yang et~al.(2019)Yang, Luo, Chen, Li, Yin, He, and Sun}]{yang2019knowledgeable}
Pengcheng Yang, Fuli Luo, Peng Chen, Lei Li, Zhiyi Yin, Xiaodong He, and Xu~Sun. 2019.
\newblock Knowledgeable storyteller: A commonsense-driven generative model for visual storytelling.
\newblock In \emph{IJCAI}, volume~3, page~7.

\bibitem[{Yang et~al.(2018)Yang, Tang, Zhang, and Cai}]{yang2018autoencodingscenegraphsimage}
Xu~Yang, Kaihua Tang, Hanwang Zhang, and Jianfei Cai. 2018.
\newblock \href {http://arxiv.org/abs/1812.02378} {Auto-encoding scene graphs for image captioning}.

\bibitem[{Yao et~al.(2019)Yao, Peng, Weischedel, Knight, Zhao, and Yan}]{yao2019planandwritebetterautomaticstorytelling}
Lili Yao, Nanyun Peng, Ralph Weischedel, Kevin Knight, Dongyan Zhao, and Rui Yan. 2019.
\newblock \href {http://arxiv.org/abs/1811.05701} {Plan-and-write: Towards better automatic storytelling}.

\bibitem[{Yao et~al.(2018)Yao, Pan, Li, and Mei}]{yao2018exploringvisualrelationshipimage}
Ting Yao, Yingwei Pan, Yehao Li, and Tao Mei. 2018.
\newblock \href {http://arxiv.org/abs/1809.07041} {Exploring visual relationship for image captioning}.

\bibitem[{Yu et~al.(2017)Yu, Bansal, and Berg}]{yu2017hierarchicallyattentivernnalbumsummarization}
Licheng Yu, Mohit Bansal, and Tamara~L. Berg. 2017.
\newblock \href {http://arxiv.org/abs/1708.02977} {Hierarchically-attentive rnn for album summarization and storytelling}.

\bibitem[{Yu et~al.(2021)Yu, Chung, Yun, Kim, and Kim}]{Yu_2021_CVPR}
Youngjae Yu, Jiwan Chung, Heeseung Yun, Jongseok Kim, and Gunhee Kim. 2021.
\newblock Transitional adaptation of pretrained models for visual storytelling.
\newblock In \emph{Proceedings of the IEEE/CVF Conference on Computer Vision and Pattern Recognition (CVPR)}, pages 12658--12668.

\bibitem[{Zhang et~al.(2024)Zhang, Huang, Jin, and Lu}]{zhang2024visionlanguagemodelsvisiontasks}
Jingyi Zhang, Jiaxing Huang, Sheng Jin, and Shijian Lu. 2024.
\newblock \href {http://arxiv.org/abs/2304.00685} {Vision-language models for vision tasks: A survey}.

\end{thebibliography}

\end{document}